\documentclass[10pt,journal,compsoc]{IEEEtran}
\pdfoutput=1
\usepackage{cite}
\usepackage{graphicx}
\usepackage{balance}  
\usepackage{subcaption}

\usepackage{booktabs} 
\usepackage[ruled,lined,linesnumbered]{algorithm2e}
\usepackage{color}
\usepackage[hyphens]{url}
\usepackage{amsfonts}
\usepackage{array,multirow}
 
\SetAlFnt{\small}
\SetAlCapFnt{\small}
\SetAlCapNameFnt{\small}
\SetAlCapHSkip{0pt}
\IncMargin{-\parindent}

\usepackage{amsmath}

\newcommand{\customparagraph}[1] {\\[6pt] \noindent {\em #1}}

\begin{document}
\title{A Survey on Data Collection for Machine~Learning\\ \large A Big Data - AI Integration Perspective}




\author{Yuji~Roh,
        Geon~Heo,
        Steven Euijong Whang,~\IEEEmembership{Senior Member,~IEEE}
\IEEEcompsocitemizethanks{\IEEEcompsocthanksitem Y. Roh, G. Heo, and S. E. Whang are with the School of Electrical Engineering, KAIST, Daejeon, Korea.\protect\\
E-mail: \{yuji.roh, geon.heo, swhang\}@kaist.ac.kr
\IEEEcompsocthanksitem Corresponding author: S. E. Whang}
}



\IEEEtitleabstractindextext{%
\begin{abstract}
Data collection is a major bottleneck in machine learning and an active research topic in multiple communities. There are largely two reasons data collection has recently become a critical issue. First, as machine learning is becoming more widely-used, we are seeing new applications that do not necessarily have enough labeled data. Second, unlike traditional machine learning, deep learning techniques automatically generate features, which saves feature engineering costs, but in return may require larger amounts of labeled data. Interestingly, recent research in data collection comes not only from the machine learning, natural language, and computer vision communities, but also from the data management community due to the importance of handling large amounts of data. In this survey, we perform a comprehensive study of data collection from a data management point of view. Data collection largely consists of data acquisition, data labeling, and improvement of existing data or models. We provide a research landscape of these operations, provide guidelines on which technique to use when, and identify interesting research challenges. The integration of machine learning and data management for data collection is part of a larger trend of Big data and Artificial Intelligence (AI) integration and opens many opportunities for new research.

\end{abstract}

}

\maketitle
\IEEEdisplaynontitleabstractindextext
\IEEEpeerreviewmaketitle

\IEEEraisesectionheading{\section{Introduction}\label{sec:introduction}}

\begin{figure*}[t]
\center
  \includegraphics[width=0.7\textwidth]{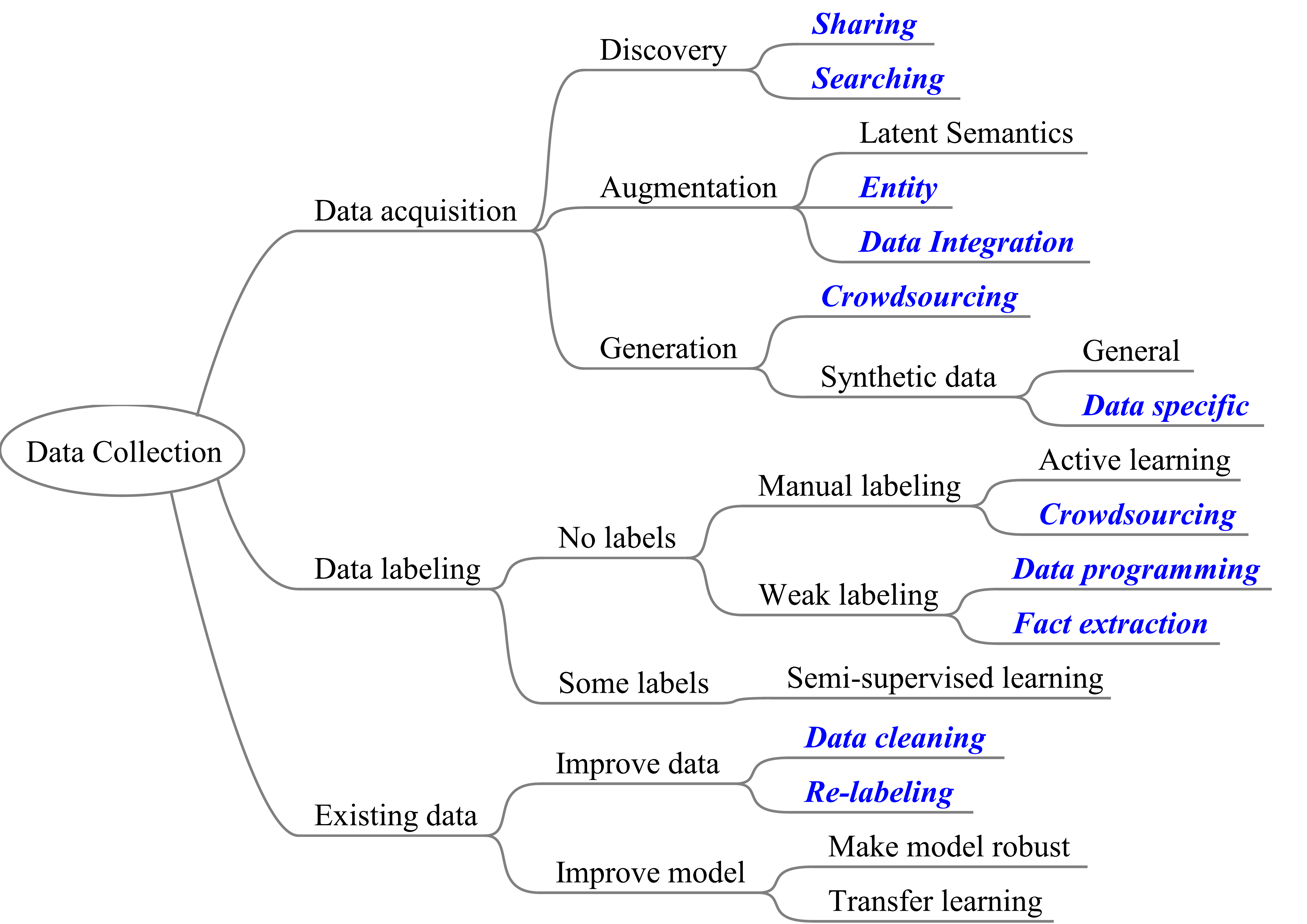}
  \caption{A high level research landscape of data collection for machine learning. The topics that are at least partially contributed by the data management community are highlighted using blue italic text. Hence, to fully understand the research landscape, one needs to look at the literature from the viewpoints of both the machine learning and data management communities.}
  \label{fig:landscape}
\end{figure*}

\IEEEPARstart{W}{e} are living in exciting times where machine learning is having a profound influence on a wide range of applications from text understanding, image and speech recognition, to health care and genomics. As a striking example, deep learning techniques are known to perform on par with ophthalmologists on identifying diabetic eye diseases in images~\cite{diabetic}. Much of the recent success is due to better computation infrastructure and large amounts of training data.

Among the many challenges in machine learning, data collection is becoming one of the critical bottlenecks. It is known that the majority of the time for running machine learning end-to-end is spent on preparing the data, which includes collecting, cleaning, analyzing, visualizing, and feature engineering. While all of these steps are time-consuming, data collection has recently become a challenge due to the following reasons.

First, as machine learning is used in new applications, it is usually the case that there is not enough training data. Traditional applications like machine translation or object detection enjoy massive amounts of training data that have been accumulated for decades. On the other hand, more recent applications have little or no training data. As an illustration, smart factories are increasingly becoming automated where product quality control is performed with machine learning. Whenever there is a new product or a new defect to detect, there is little or no training data to start with. The na\"ive approach of manual labeling may not be feasible because it is expensive and requires domain expertise. This problem applies to any novel application that benefits from machine learning.

Moreover, as deep learning~\cite{Goodfellow:2016:DL:3086952} becomes popular, there is even more need for training data. In traditional machine learning, feature engineering is one of the most challenging steps where the user needs to understand the application and provide features used for training models. Deep learning, on the other hand, can automatically generate features, which saves us of feature engineering, which is a significant part of data preparation. However, in return, deep learning may require larger amounts of training data to perform well~\cite{DBLP:conf/icml/BachHRR17}.

As a result, there is a pressing need of accurate and scalable data collection techniques in the era of Big data, which motivates us to conduct a comprehensive survey of the data collection literature from a data management point of view. There are largely three methods for data collection. First, if the goal is to share and search new datasets, then data acquisition techniques can be used to discover, augment, or generate datasets. Second, once the datasets are available, various data labeling techniques can be used to label the individual examples. Finally, instead of labeling new datasets, it may be better to improve existing data or train on top of trained models. These three methods are not necessarily distinct and can be used together. For example, one could search and label more datasets while improving existing ones.

An interesting observation is that the data collection techniques come not only from the machine learning community (including natural language processing and computer vision, which traditionally use machine learning heavily), but have also been studied for decades by the data management community, mainly under the names of data science and data analytics. Figure~\ref{fig:landscape} shows an overview of the research landscape where the topics that have contributions from the data management community are highlighted with blue italic text. Traditionally, labeling data has been a natural focus of research for machine learning tasks. For example, semi-supervised learning is a classical problem where model training is done on a small amount of labeled data and a larger amount of unlabeled data. However, as machine learning needs to be performed on large amounts of training data, data management issues including how to acquire large datasets, how to perform data labeling at scale, and how to improve the quality of large amounts of existing data become more relevant. Hence, to fully understand the research landscape of data collection, one needs to understand the literature from both the machine learning and data management communities.

\begin{figure}[t]
\center
  \includegraphics[width=0.48\textwidth]{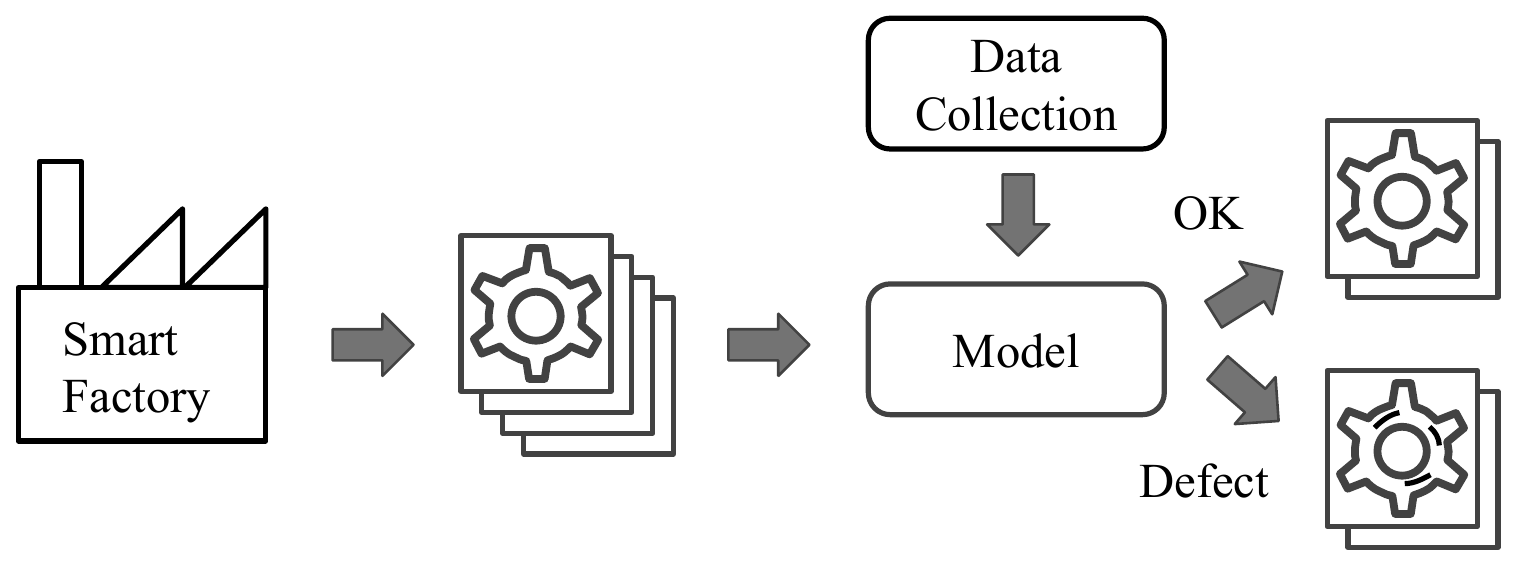}
  \caption{A running example for data collection. A smart factory may produce various images of product components, which are classified as normal or defective by a convolutional neural network model. Unfortunately, with an application this specific, it is often difficult to find enough data for training the model.}
  \label{fig:scenario}
\end{figure}

While there are many surveys on data collection that are either limited to one discipline or a class of techniques, to our knowledge, this survey is the first to bridge the machine learning (including natural language processing and computer vision) and data management disciplines. We contend that a machine learning user needs to know the techniques on all sides to make informed decisions on which techniques to use when. In fact, data management plays a role in almost all aspects of machine learning~\cite{polyzotis2018data,Polyzotis:2017:DMC:3035918.3054782}. We note that many sub-topics including semi-supervised learning, active learning, and transfer learning are large enough to have their own surveys. The goal of this survey is not to go into all the depths of these sub-topics, but to focus on breadth and identify what data collection techniques are relevant for machine learning purposes and what research challenges exist. Hence, we will only cover the most representative work of the sub-topics, which are either the best-performing or most recent ones. The key audience of this survey can be researchers or practitioners that are starting to use data collection for machine learning and need an overall landscape introduction. Since the data collection techniques come from different disciplines, some may involve relational data while others non-relational data (e.g., images and text). Sometimes the boundary between operations (e.g., data acquisition and data labeling) is not clear cut. In those cases, we will clarify that the techniques are relevant in multiple operations.
\begin{figure*}[t]
\center
  \includegraphics[width=\textwidth]{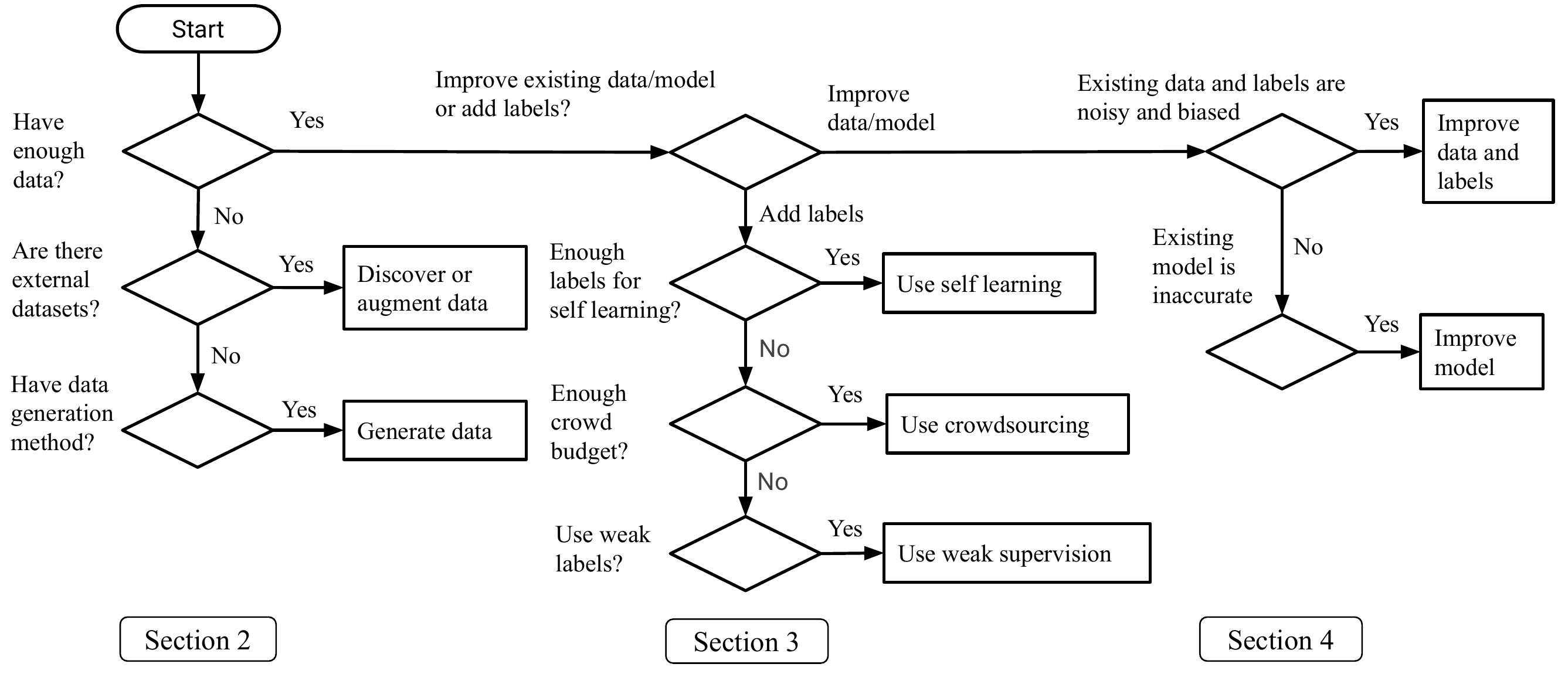}
  \caption{A decision flow chart for data collection. From the top left, Sally can start by asking whether she has enough data. The following questions lead to specific techniques that can be used for acquiring data, labeling data, or improving existing data or models. This flow chart does not cover all the details in this survey. For example, data labeling techniques like self learning and crowdsourcing can be performed together as described in Section~\ref{sec:activelearning}. Also, some questions (e.g., ``Enough labels for self learning?'') are not easy to answer and may require an in-depth understanding of the application and data. There are also techniques specific to the data type (images and text), which we detail in the body of the paper.}
  \label{fig:workflow}
\end{figure*}
\customparagraph{Motivating Example}
To motivate the need to explore the techniques in Figure~\ref{fig:landscape}, we present a running example on data collection based on our experience with collaborating with the industry on a smart factory application. Suppose that Sally is a data scientist who works on product quality for a smart factory. The factory may produce manufacturing components like gears where it is important for them not to have scratches, dents, or any foreign substance. Sally may want to train a model on images of the components, which can be used to automatically classify whether each product has defects or not. This application scenario is depicted in Figure~\ref{fig:scenario}. A general decision flow chart of the data collection techniques that Sally can use is shown in Figure~\ref{fig:workflow}. Although the chart may look complicated at first glance, we contend that it is necessary to understand the entire research landscape to make informed decisions for data collection. In comparison, recent commercial tools~\cite{googlecloudautoml,microsoftcustomvision,sagemaker} only cover a subset of all the possible data collection techniques. When using the chart, one can quickly narrow down the options in two steps by deciding whether to perform one of data acquisition, data labeling, or existing data improvements, and then choosing the specific technique to use for each operation. For example, if there is no data, then Sally could generate a dataset by installing camera equipment. Then if she has enough budget for human computation, she can use crowdsourcing platforms like Amazon Mechanical Turk to label the product images for defects. We will discuss more details of the flow chart in the following sections.

The rest of the paper is organized as follows:

\begin{itemize}
    \item We review the data acquisition literature, which can be categorized into data discovery, data augmentation, and data generation. Many of the techniques require scalable solutions and have thus been studied by the data management community (Section~\ref{sec:data-acquisition}).
    \item We review the data labeling literature and group the techniques into three approaches: utilizing existing labels, using crowdsourcing techniques, and using weak supervision. While data labeling is traditionally a machine learning topic, it is also studied in the data management community as scalability becomes an issue (Section~\ref{sec:data-labeling}).
    \item We review techniques for improving existing data or models when acquiring and labeling new data is not the best option. Improving data quality through cleaning is a traditional data management topic where recent techniques are increasingly focusing on machine learning applications (Section~\ref{sec:without-data}).
    \item We put all the techniques together and provide guidelines on how to decide which data collection techniques to use when (Section~\ref{sec:which-technique}).
    \item Based on the current research landscape, we identify interesting future research challenges (Section~\ref{sec:future-work}).
\end{itemize}

\section{Data Acquisition}
\label{sec:data-acquisition}

\begin{table*}[t]
\renewcommand{\arraystretch}{1.3}
\centering
\begin{tabular}{ccccc}
\hline
{\bf Task}  & {\bf Approach} & {\bf Techniques} \\
\hline \hline
\multirow{5}{*}{Data discovery} & \multirow{3}{*}{Sharing}  & Collaborative Analysis~\cite{Bhardwaj:2015:CDA:2824032.2824100,DBLP:conf/cidr/BhardwajBCDEMP15,Bhattacherjee:2015:PDV:2824032.2824035} \\
 & & Web~\cite{DBLP:conf/cidr/Halevy13,DBLP:conf/cloud/GonzalezHJLMSS10,DBLP:conf/sigmod/GonzalezHJLMSSG10,ckan,quandl,datamarket} \\
 & & Collaborative and Web~\cite{kaggle}  \\\cline{2-3}
  & \multirow{2}{*}{Searching} & Data Lake~\cite{DBLP:conf/cidr/TerrizzanoSRC15, Halevy:2016:GOG:2882903.2903730,CastroFernandez:2017:DDC:3035918.3058740,DBLP:conf/cidr/DengFAWSEIMO017,DBLP:conf/sigmod/GaoHP18} \\
 & & Web~\cite{Cafarella:2008:WEP:1453856.1453916, DBLP:journals/pvldb/CafarellaHLMYWW18,datasetsearch,Elmeleegy:2011:HRT:1969331.1969352,Chu:2015:TTE:2723372.2723725,DBLP:journals/debu/ChakrabartiCCGH16,Cafarella:2009:DIR:1687627.1687750,Yakout:2012:IEA:2213836.2213848,DBLP:journals/pvldb/DalviKS11,DBLP:conf/sigmod/BohannonDFJKK12,DBLP:reference/db/BaumgartnerGG18}\\\hline
\multirow{3}{*}{Data augmentation} & &  Deriving Latent Semantics~\cite{DBLP:conf/nips/MikolovSCCD13,pennington2014glove,DBLP:journals/jmlr/BleiNJ03}\\
 & & \multirow{1}{*}{Entity Augmentation}~\cite{Cafarella:2009:DIR:1687627.1687750,Yakout:2012:IEA:2213836.2213848} \\
 & & Data Integration~\cite{Kumar:2016:JJT:2882903.2882952,Shah:2017:KKJ:3173074.3173084,DBLP:journals/debu/StonebrakerI18,DBLP:books/daglib/0029346,DBLP:conf/sigmod/LiC019,DBLP:journals/pvldb/ChenKNP17,DBLP:conf/sigmod/KumarNPZ16} \\\hline
\multirow{6}{*}{Data generation} & \multirow{2}{*}{Crowdsourcing} & Gathering~\cite{Little:2010:THC:1866029.1866040,Barowy:2012:API:2384616.2384663,Ahmad:2011:JPE:2047196.2047203,DBLP:journals/pvldb/ParkPPGPW12,Franklin:2011:CAQ:1989323.1989331,DBLP:conf/cidr/MarcusWMM11,DBLP:conf/icde/BoimGMNPT12,Franklin:2013:CEQ:2510649.2511160,DBLP:conf/sigmod/ParkW14,alfredo:crowdsource} \\
 & & Processing~\cite{Franklin:2011:CAQ:1989323.1989331,DBLP:conf/cidr/StonebrakerBIBCZPX13,Gokhale:2014:CHC:2588555.2588576,DBLP:conf/cidr/MarcusWMM11} \\\cline{2-3}
& \multirow{4}{*}{Synthetic Data} & Generative Adversarial Networks~\cite{Goodfellow:2014:GAN:2969033.2969125,DBLP:conf/mlhc/ChoiBMDSS17,DBLP:journals/pvldb/ParkMGJPK18,DBLP:journals/corr/abs-1811-11264,DBLP:journals/corr/Goodfellow17,DBLP:journals/corr/abs-1807-04720} \\
& & Policies~\cite{DBLP:journals/corr/abs-1805-09501,DBLP:conf/nips/RatnerEHDR17}
\\
& & Image~\cite{Peng:2015:LDO:2919332.2920015,DBLP:conf/icra/KimCAK17,DBLP:journals/corr/abs-1804-02684,DBLP:journals/corr/JaderbergSVZ14, Gupta_2016_CVPR,DBLP:conf/eccv/XiaCWS14,DBLP:conf/mm/BaiYYXMZ15} \\
& & Text~\cite{E17-1083,DBLP:journals/corr/abs-1804-06059,sears:acl18} \\\hline
\end{tabular}
\caption{A classification of data acquisition techniques. Some of the techniques can be used together. For example, data can be generated while augmenting existing data.}
\label{tbl:data_acquisition}
\end{table*}

The goal of data acquisition is to find datasets that can be used to train machine learning models. There are largely three approaches in the literature: data discovery, data augmentation, and data generation. Data discovery is necessary when one wants to share or search for new datasets and has become important as more datasets are available on the Web and corporate data lakes~\cite{DBLP:journals/debu/HalevyKNOPRW16,DBLP:conf/cidr/TerrizzanoSRC15}. Data augmentation complements data discovery where existing datasets are enhanced by adding more external data. Data generation can be used when there is no available external dataset, but it is possible to generate crowdsourced or synthetic datasets instead. The following sections will cover the three operations in more detail. The individual techniques are classified in Table~\ref{tbl:data_acquisition}.

\subsection{Data Discovery}

Data discovery can be viewed as two steps. First, the generated data must be indexed and published for sharing. Many collaborative systems are designed to make this process easy. However, other systems are built without the intention of sharing datasets. For these systems, a {\em post-hoc} approach must be used where metadata is generated after the datasets are created, without the help of the dataset owners. Next, someone else can search the datasets for their machine learning tasks. Here the key challenges include how to scale the searching and how to tell whether a dataset is suitable for a given machine learning task.  While most of the data discovery literature came from the data management community for data science and data analytics, they are also relevant in a machine learning context. However, another challenge in machine learning is data labeling, which we cover in Section~\ref{sec:data-labeling}.


\subsubsection{Data Sharing}

We study data systems that are designed with dataset sharing in mind. These systems may focus on collaborative analysis, publishing on the Web, or both. 
\customparagraph{Collaborative Analysis} In an environment where data scientists are collaboratively analyzing different versions of datasets, DataHub~\cite{Bhardwaj:2015:CDA:2824032.2824100,DBLP:conf/cidr/BhardwajBCDEMP15,Bhattacherjee:2015:PDV:2824032.2824035} can be used to host, share, combine, and analyze them. There are two components: a dataset version control system inspired by Git (a version control system for code) and a hosted platform on top of it, which provides data search, data cleaning, data integration, and data visualization. A common use case of DataHub is where individuals or teams run machine learning tasks on their own versions of a dataset and later merge with other versions if necessary.
\customparagraph{Web} A different approach of sharing datasets is to publish them on the Web. Google Fusion Tables~\cite{DBLP:conf/cidr/Halevy13,DBLP:conf/cloud/GonzalezHJLMSS10,DBLP:conf/sigmod/GonzalezHJLMSSG10} is a cloud-based service for data management and integration. Fusion Tables enables users to upload structured data (e.g., spreadsheets) and provides tools for visually analyzing, filtering, and aggregating the data. The datasets that are published through Fusion Tables on the Web can be crawled by search engines and show up in search results. The datasets are therefore primarily accessible through Web search. Fusion Tables has been widely used in data journalism for creating interactive maps of data and adding them in articles. In addition, there are many data marketplaces including CKAN~\cite{ckan}, Quandl~\cite{quandl}, and DataMarket~\cite{datamarket} where users can buy and sell datasets or find public datasets.
\customparagraph{Collaborative and Web} More recently, we are seeing a merging of collaborative and Web-based systems. For example, Kaggle~\cite{kaggle} makes it easy to share datasets on the Web and even host data science competitions for models trained on the datasets. A Kaggle competition host posts a dataset along with a description of the challenge. Participants can then experiment with their techniques and compete with each other. After the deadline passes, a prize is given to the winner of the competition. Kaggle currently has thousands of public datasets and code snippets (called kernels) from competitions. In comparison to DataHub and Fusion Tables, the Kaggle datasets are coupled with competitions and are thus more readily usable for machine learning purposes.

\subsubsection{Data Searching}

While the previous data systems are platforms for sharing datasets, as a next logical step, we now explore systems that are mainly designed for searching datasets. This setting is common within large companies or on the Web.
\customparagraph{Data Lake } Data searching systems have become more popular with the advent of {\em data lakes}~\cite{DBLP:journals/debu/HalevyKNOPRW16,DBLP:conf/cidr/TerrizzanoSRC15} in corporate environments where many datasets are generated internally, but they are not easily discoverable by other teams or individuals within the company. Providing a way to search datasets and analyze them has significant business value because the teams or individuals do not have to make redundant efforts to re-generate the datasets for their machine learning tasks. Most of the recent data lake systems have come from the industry. In many cases, it is not feasible for all the dataset owners to publish datasets through one system. Instead, a post-hoc approach becomes necessary where datasets are processed for searching after they are created, and no effort is required on the dataset owner's side.

As an early solution for data lakes, IBM proposed a system~\cite{DBLP:conf/cidr/TerrizzanoSRC15} that enables datasets to be curated and then searched. IBM estimates that 70\% of the time spent on analytic projects is concerned with discovering, cleaning, and integrating datasets that are scattered among many business applications. Thus, IBM takes the stance of creating, filling, maintaining, and governing the data lake where these processes are collectively called {\em data wrangling}. When analyzing data, users do not perform the analytics directly on the data lake, but extract data sets and store them separately. Before this step, the users can do a preliminary exploration of datasets, e.g., visualizing them to determine if the dataset is useful and does not contain anomalies that need further investigation. While supporting data curation in the data lake saves users from processing raw data, it does limit the scalability of how many datasets can be indexed.

More recently, scalability has become a pressing issue for handling data lakes that consists of most datasets in a large company. Google Data Search ({\sc Goods})~\cite{Halevy:2016:GOG:2882903.2903730} is a system that catalogues the metadata of tens of billions of datasets from various storage systems within Google. {\sc Goods} infers various metadata including owner information and provenance information (by looking up job logs), analyzes the contents of the datasets, and collects input from users. At the core is a central catalog, which contains the metadata and is indexed for data searching. Due to Google's scale, there are many technical challenges including scaling to the number of datasets, supporting a variety of data formats where the costs for extracting metadata may differ, updating the catalog entries due to the frequent churn of datasets, dealing with uncertainty in metadata discovery, computing dataset importance for search ranking, and recovering dataset semantics that are missing. To find datasets, users can use keywords queries on the {\sc Goods} frontend and view profile pages of the datasets that appear in the search results. In addition, users can track the provenance of a dataset to see which datasets were used to create the given dataset and those that rely on it.

Finally, expressive queries are also important for searching a data lake. While {\sc Goods} scales, one downside is that it only supports simple keyword queries. This approach is similar to keyword search in databases~\cite{DBLP:journals/debu/YuQC10,DBLP:journals/pvldb/ChaudhuriD09}, but the purpose is to find datasets instead of tuples. The {\sc Data Civilizer} system~\cite{CastroFernandez:2017:DDC:3035918.3058740,DBLP:conf/cidr/DengFAWSEIMO017} complements {\sc Goods} by focusing more on the discovery aspect of datasets. Specifically, {\sc Data Civilizer} consists of a module for building a linkage graph of data. Assuming that datasets have schema, the nodes in the linkage graph are columns of tables while edges are relationships like primary key-foreign key (PK-FK) relationships. A data discovery module then supports a rich set of {\em discovery queries} on the linkage graph, which can help users more easily discover the relevant datasets. {\sc Dataraman}~\cite{DBLP:conf/sigmod/GaoHP18} specializes in extracting structured data from semi-structured log datasets in data lakes automatically by learning patterns. {\sc Aurum}~\cite{DBLP:conf/icde/FernandezAKYMS18,DBLP:conf/icde/FernandezMQEIMO18} supports data discovery queries on semantically-linked datasets.
\customparagraph{Web }
As the Web contains large numbers of structured datasets, there have been significant efforts to automatically extract the useful ones~\cite{DBLP:journals/pvldb/DalviKS11,DBLP:conf/sigmod/BohannonDFJKK12,DBLP:reference/db/BaumgartnerGG18}. One of the most successful systems is WebTables~\cite{DBLP:journals/pvldb/CafarellaHLMYWW18,Cafarella:2008:WEP:1453856.1453916}, which automatically extracts structured data that is published online in the form of HTML tables. For example, WebTables extracts all Wikipedia infoboxes. Initially, about 14.1 billion HTML tables are collected from the Google search web crawl. Then a classifier is applied to determine which tables can be viewed as relational database tables. Each relational table consists of a schema that describes the columns and a set of tuples. In comparison to the above data lake systems, WebTables collects structured data from the Web.

As Web data tends to be much more diverse than say those in a corporate environment, the table extraction techniques have been extended in multiple ways as well. One direction is to extend table extraction beyond identifying HTML tags by extracting relational data in the form of vertical tables and lists and leveraging knowledge bases~\cite{Elmeleegy:2011:HRT:1969331.1969352,Chu:2015:TTE:2723372.2723725}. Table searching also evolved where, in addition to keyword searching, row-subset queries, entity-attribute queries, and column search were introduced~\cite{DBLP:journals/debu/ChakrabartiCCGH16}. Finally, techniques for enhancing the tables~\cite{Cafarella:2009:DIR:1687627.1687750,Yakout:2012:IEA:2213836.2213848} were proposed where entities or attribute values are added to make the tables more complete.

Recently, a service called Google Dataset Search~\cite{datasetsearch} was launched for searching repositories of datasets on the Web. The motivation is that there are thousands of data repositories on the Web that contain millions of datasets that are not easy to search. Dataset Search lets dataset providers describe their datasets using various metadata (e.g., author, publication date, how the data was collected, and terms for using the data) so that they become more searcheable. In comparison to the fully-automatic WebTables, dataset providers may need to do some manual work, but have the opportunity to make their datasets more searcheable. In comparison to {\sc Goods}, Dataset Search targets the Web instead of a data lake.

\subsection{Data Augmentation}

Another approach to acquiring data is to augment existing datasets with external data. In the machine learning community, adding pre-trained embeddings is a common way to increase the features to train on. In the data management community, entity augmentation techniques have been proposed to further enrich existing entity information. Data integration is a broad topic and can be considered as data augmentation if we are extending existing datasets with newly-acquired ones. 


\subsubsection{Deriving Latent Semantics}

A common data augmentation is to derive latent semantics from data. A popular technique is to generate and use embeddings that represent words, entities, or knowledge. In particular, word embeddings have been successfully used to solve many problems in natural language processing (NLP). Word2vec~\cite{DBLP:conf/nips/MikolovSCCD13} is a seminal work where, given a text corpus, a word is represented by a vector of real numbers that captures the linguistic context of the word in the corpus. The word vectors can be generated by training a shallow two-layer neural network to reconstruct the surrounding words in the corpus. There are two possible models for training word vectors: Continuous Bag-of-Words (CBOW) and Skip-gram. While CBOW predicts a word based on its surrounding words, Skip-gram does the opposite and predicts the surrounding words based on a given word. As a result, two words that occur in similar contexts tend to have similar word vectors. A fascinating application of word vectors is performing arithmetic operations on the word vectors. For example, the result of subtracting the word vector of ``king'' by that of ``queen'' is similar to the result of subtracting the word vector of ``man'' by that of ``woman''. Since word2vec was proposed, there have been many extensions including GloVe~\cite{pennington2014glove}, which improves word vectors by also taking into account global corpus statistics, and Doc2Vec~\cite{DBLP:conf/icml/LeM14}, which generates representations of documents.

Another technique for deriving latent semantics is latent topic modeling. For example, Latent Dirichlet Allocation~\cite{DBLP:journals/jmlr/BleiNJ03} (LDA) is a generative model that can be used to explain why certain parts of the data are similar using unobserved groups.

\subsubsection{Entity Augmentation}

In many cases, datasets are incomplete and need to be filled in by gathering more information. The missing information can either be values or entire features. An early system called Octopus~\cite{Cafarella:2009:DIR:1687627.1687750} composes Web search queries using keys of the table containing the entities. Then all the Web tables in the resulting Web pages are clustered by schema, and the tables in the most relevant cluster are joined with the entity table. More recently, InfoGather~\cite{Yakout:2012:IEA:2213836.2213848} takes a holistic approach using Web tables on the Web. The entity augmentation is performed by filling in missing values of attributes in some or all of the entities by matching multiple Web tables using schema matching. To help the user decide which attributes to fill in, InfoGather identifies synonymous attributes in the Web tables.

\subsubsection{Data Integration}

Data integration can also be considered as data augmentation, especially if we are extending existing data sets with other acquired ones. Since this discipline is well established, we point the readers to some excellent surveys~\cite{DBLP:journals/debu/StonebrakerI18,DBLP:books/daglib/0029346}. More recently, an interesting line of work relevant to machine learning~\cite{DBLP:conf/sigmod/LiC019,DBLP:journals/pvldb/ChenKNP17,DBLP:conf/sigmod/KumarNPZ16} observes that in practice, many companies use relational databases where the training data is divided into smaller tables. However, most machine learning toolkits assume that a training dataset is a single file and ignore the fact that there are typically multiple tables in a database due to normalization. The key question is whether joining the tables and augmenting the information is useful for model training. The Hamlet  system~\cite{Kumar:2016:JJT:2882903.2882952} and its subsequent Hamlet++ systems~\cite{Shah:2017:KKJ:3173074.3173084} address this problem by determining if key-foreign key (KFK) joins are necessary for improving the model accuracy for various classifiers (linear, decision trees, non-linear SVMs, and artificial neural networks) and propose decision rules to predict when it is safe to avoid joins and, as a result, significantly reduce the total runtime. A surprising result is that joins can often be avoided without negatively influencing the model's accuracy. Intuitively, a foreign key determines the entire record of the joining table, so the features brought in by a join do not add a lot more information. 


\subsection{Data Generation}

If there are no existing datasets that can be used for training, then another option is to generate the datasets either manually or automatically. For manual construction, crowdsourcing is the standard method where human workers are given tasks to gather the necessary bits of data that collectively become the generated dataset. Alternatively, automatic techniques can be used to generate synthetic datasets. Note that data generation can also be viewed as data augmentation if there is existing data where some missing parts needs to be filled in.

\subsubsection{Crowdsourcing}

Crowdsourcing is used to solve a wide range of problems, and there are many surveys as well~\cite{DBS-044,6488672,Daniel:2018:QCC:3177787.3148148,7420720}. One of the earliest and most popular platforms is Amazon Mechanical Turk~\cite{mturk} where tasks (called {\em HITs}) are assigned to human workers, and workers are compensated for finishing the tasks. Since then, many other crowdsourcing platforms have been developed, and research on crowdsourcing has flourished in the areas of data management, machine learning, and human computer interaction. There is a wide range of crowdsourcing tasks from simple ones like labeling images up to complex ones like collaboratively writing that involve multiple steps~\cite{DBLP:conf/cscw/KimSCB17,DBLP:conf/cscw/SalehiTIK17}. Another important usage of crowdsourcing is data labeling (e.g., the ImageNet project), which we discuss in Section~\ref{sec:crowdbasedtechniques}.

In this section, we narrow the scope and focus on crowdsourcing techniques that are specific to data generation tasks. A recent survey~\cite{Garcia-Molina:2016:CDC:2914178.2914228} provides an extensive discussion on the challenges for data crowdsourcing. Another survey~\cite{Amsterdamer:2015:FCD:2737817.2737819} touches on the theoretical foundations of data crowdsourcing. According to both surveys, data generation using crowdsourcing can be divided into two steps: {\em gathering} data and {\em preprocessing} data. 
\customparagraph{Gathering data} One way to categorize data gathering techniques is whether the tasks are procedural or declarative. A procedural task is where the task creator defines explicit steps and assigns them to workers. For example, one may write a computer program that issues tasks to workers. TurKit~\cite{Little:2010:THC:1866029.1866040} allows users to write scripts that include HITs using a crash-and-return programming model where a script can be re-executed without re-running costly functions with side effects. {\sc AutoMan}~\cite{Barowy:2012:API:2384616.2384663} is a domain-specific language embedded in Scala where crowdsourcing tasks can be invoked like conventional functions. {\sc Dog}~\cite{Ahmad:2011:JPE:2047196.2047203} is a high-level programming language that compiles into MapReduce tasks that can be performed by humans or machines. A declarative task is when the task creator specifies high-level data requirements, and the workers provide the data that satisfy them. For example, a database users may pose an SQL query like ``SELECT title, director, genre, rating FROM MOVIES WHERE genre = 'action''' to gather movie ratings data for a recommendation system. {\sc Deco}~\cite{DBLP:journals/pvldb/ParkPPGPW12} uses a simple extension of SQL and defines precise semantics for arbitrary queries on stored data and data collected by the crowd. CrowdDB~\cite{Franklin:2011:CAQ:1989323.1989331} focuses on the systems aspect of using crowdsourcing to answer queries that cannot be answered automatically.

Another way to categorize data gathering is whether the data is assumed to be closed-world or open-world. Under a closed-world assumption, the data is assumed to be ``known'' and entirely collectable by asking the right questions. {\sc AskIt!}~\cite{DBLP:conf/icde/BoimGMNPT12} uses this assumption and focuses on the problem of determining which questions should be directed to which users, in order to minimize the uncertainty of the collected data. In an open-world assumption, there is no longer a guarantee that all the data can be collected. Instead, one must estimate if enough data was collected. 
Statistical tools~\cite{Franklin:2013:CEQ:2510649.2511160} have been proposed for scanning a single table with predicates like ``SELECT FLAVORS FROM ICE\_CREAM.'' Initially, many flavors can be collected, but the rate of new flavors will inevitably slow down, and statistical methods are used to estimate the future rate of new values. 

Data gathering is not limited to collecting entire records of a table. CrowdFill~\cite{DBLP:conf/sigmod/ParkW14} is a system for collecting parts of structured data from the crowd. Instead of posing specific questions to workers, CrowdFill shows a partially-filled table. Workers can then fill in the empty cells and also upvote or downvote data entered by other workers. CrowdFill provides a collaborative environment and allows the specification of constraints on values and mechanisms for resolving conflicts when workers are filling in values of the same record. {\sc ALFRED}~\cite{alfredo:crowdsource} uses the crowd to train extractors that can then be used to acquire data. {\sc ALFRED} asks simple yes/no membership questions on the contents of Web pages to workers and uses the answers to infer the extraction rules. The quality of the rules can be improved by recruiting multiple workers.
\customparagraph{Preprocessing data } Once the data is gathered, one may want to preprocess the data to make it suitable for machine learning purposes. While many possible crowd operations have been proposed, the ones that are relevant include data curation, entity resolution, and joining datasets. Data Tamer~\cite{DBLP:conf/cidr/StonebrakerBIBCZPX13} is an end-to-end data curation system that can clean and transform datasets and semantically integrate with other datasets. Data Tamer has a crowdsourcing component (called Data Tamer Exchange), which assigns tasks to workers. The supported operations are attribute identification (i.e., determine if two attributes are the same) and entity resolution (i.e., determine if two entities are the same). Corleone~\cite{Gokhale:2014:CHC:2588555.2588576} is a hands-off crowdsourcing system, which crowdsources the entire workflow of entity resolution to workers. CrowdDB~\cite{Franklin:2011:CAQ:1989323.1989331} and Qurk~\cite{DBLP:conf/cidr/MarcusWMM11} are systems for aggregating, sorting, and joining datasets. 

For both gathering and preprocessing data, quality control is an important challenge as well. The issues include designing the right interface to maximize worker productivity, managing workers who may have different levels of skills (or may even be spammers), and decomposing problems into smaller tasks and aggregating them. Several surveys~\cite{6488672,Daniel:2018:QCC:3177787.3148148,7420720} cover these issues in detail.

\subsubsection{Synthetic Data Generation}

Generating synthetic data along with labels is increasingly being used in machine learning due to its low cost and flexibility~\cite{DBLP:conf/dsaa/PatkiWV16}. A simple method is to start from a probability distribution and generate a sample from that distribution using tools like scikit learn~\cite{Pedregosa:2011:SML:1953048.2078195}. In addition, there are more advanced techniques like Generative Adversarial Networks (GANs)~\cite{Goodfellow:2014:GAN:2969033.2969125,Goodfellow:2016:DL:3086952,DBLP:journals/corr/Goodfellow17,DBLP:journals/corr/abs-1807-04720} and application-specific generation techniques. We first provide a brief introduction of GANs and present synthetic data generation techniques on relational data. We then introduce recent augmentation techniques using policies. Finally, we introduce image and text data generation techniques due to their importance. 
\customparagraph{GANs} The key approach of a GAN is to train two contesting neural networks: a generative network and a discriminative network. The generative network learns to map from a latent space to a data distribution, and the discriminative network discriminates examples from the true distribution from the candidates produced by the generative network.

The training of a GAN can be formalized as:
\begin{equation*}
\min_G \max_D V(D, G)
\end{equation*}
\begin{equation*}
V(D, G) = \mathop{\mathbb{E}}_{x \sim p_{data}(x)}[\mathrm{log}D(x)] + \mathop{\mathbb{E}}_{z \sim p_z(z)}[\mathrm{log}(1-D(G(z))]
\end{equation*}
where $p_{data}(x)$ is the distribution of the real data, $p_z(z)$ is the distribution of the generator, $G(z)$ is the generative network, and $D(x)$ is the discriminative network.

The objective of the generative network is to increase the error rate of the discriminative network. That is, the generative network attempts to fool the discriminative network into thinking that its candidates are from the true distribution. GANs have been used to generate synthetic images and videos that look realistic in many applications.

GANs have recently been used to generate synthetic relational data. A {\sc medGAN}~\cite{DBLP:conf/mlhc/ChoiBMDSS17} generates synthetic patient records with high-dimensional discrete variable (binary or count) features based on real patient records. While GANs can only learn to approximate discrete patient records, the novelty is to also use an autoencoder to project these records into a lower dimensional space and then project them back to the original space. A {\sc table-GAN}~\cite{DBLP:journals/pvldb/ParkMGJPK18} also synthesizes tables that are similar to the real ones, but with a focus on privacy perservation. In particular, a metric for information loss is defined, and two parameters are provided to adjust the information loss. The higher the loss, the more privacy the synthetic table has. A {\sc TGAN}~\cite{DBLP:journals/corr/abs-1811-11264} focuses on simultaneously generating values for a mixture of discrete and continuous features. 
\customparagraph{Policies}
Another recent approach is to use human-defined policies~\cite{DBLP:journals/corr/abs-1805-09501,DBLP:conf/nips/RatnerEHDR17} to apply transformations to the images as long as they remain realistic. This criteria can be enforced by training a reinforcement learning model on a separate validation set.
\customparagraph{Data-specific} We now introduce data-specific techniques for generation. Synthetic image generation is a heavily-studied topic in the computer vision community. Given the wide range of vision problems, we are not aware of a comprehensive survey on synthetic data generation and will only focus on a few representative problems. In object detection, it is possible to learn 3D models of objects and give variations (e.g., rotate a car 90 degrees) to generate another realistic image~\cite{Peng:2015:LDO:2919332.2920015,DBLP:conf/icra/KimCAK17}. If the training data is a rapid sequence of images frames in time~\cite{DBLP:journals/corr/abs-1804-02684} the objects of a frame can be assumed to move in a linear trajectory between consecutive frames. Text within images is another application where one can vary the fonts, sizes, and colors of the text to generate large amounts of synthetic text images~\cite{DBLP:journals/corr/JaderbergSVZ14, Gupta_2016_CVPR}.

An alternative approach to generating image datasets is to start from a large set of noisy images and select the clean ones. Xia et al.~\cite{DBLP:conf/eccv/XiaCWS14} searches the Web for images with noise and then uses a density-based measure to cluster the images and remove outliers. Bai et al.~\cite{DBLP:conf/mm/BaiYYXMZ15} exploits large click-through logs, which contains queries of users and the images that were clicked by those users. A deep neural network is used to learn representations of the words and images and compute word-word and image-word similarities. The noisy images that have low similarities to their categories are then removed.

Generating synthetic text data has also been studied in the natural language processing community. Paraphrasing~\cite{E17-1083} is a classical problem of generating alternative expressions that have the same semantic meaning. For example ``What does $X$ do for a living?'' is a paraphrase of ``What is $X$'s job?''. We briefly cover two recent methods -- one is syntax-based and the other semantics-based -- that uses paraphrasing to generate large amounts of synthetic text data. Syntactically controlled paraphrase networks~\cite{DBLP:journals/corr/abs-1804-06059} (SCPNs) can be trained to produce paraphrases of a sentence with different sentence structures. Semantically equivalent adversarial rules for text~\cite{sears:acl18} (SEARs) have been proposed for perturbing input text while preserving its semantics. SEARs can be used to debug a model by applying them on training data and seeing if the re-trained model changes its predictions. In addition, there are many paraphrasing techniques that are not covered in this survey.

\section{Data Labeling}
\label{sec:data-labeling}

\begin{table*}[t]
\renewcommand{\arraystretch}{1.3}
\centering
\begin{tabular}{ccccc}
\hline
{\bf Category} & {\bf Approach} & {\bf Machine learning task} & {\bf Data types} & {\bf Techniques} \\
\hline \hline
\multirow{3}{*}{Use Existing Labels} & \multirow{2}{*}{Self-labeled} & \multirow{1}{*}{classification} & \multirow{1}{*}{all} & \cite{Yarowsky:1995:UWS:981658.981684,Zhou:2005:TEU:1092713.1092809,DBLP:conf/ictai/ZhouG04,Blum:1998:CLU:279943.279962,DBLP:journals/kais/TrigueroGH15} \\
\cline{3-5}
 & & \multirow{1}{*}{regression} & \multirow{1}{*}{all} & \cite{Brefeld:2006:ECL:1143844.1143862,Sindhwani05aco-regularized,Zhou:2005:SRC:1642293.1642439} \\\cline{2-5}
& \multirow{1}{*}{Label propagation} & \multirow{1}{*}{classification} & \multirow{1}{*}{graph} & \cite{DBLP:conf/aistats/RaviD16,Zhu:2003:SLU:3041838.3041953,DBLP:conf/aistats/TalukdarC14}  \\
\hline
\multirow{7}{*}{Crowd-based} & \multirow{2}{*}{Active learning} & \multirow{1}{*}{classification} &  \multirow{1}{*}{all} & \cite{Lewis:1994:SAT:188490.188495,Seung:1992:QC:130385.130417,DBLP:series/synthesis/2012Settles,Abe:1998:QLS:645527.657478,Settles:2008:AAL:1613715.1613855,Settles:2007:MAL:2981562.2981724,Roy:2001:TOA:645530.655646} \\\cline{3-5}
& & \multirow{1}{*}{regression} &  \multirow{1}{*}{all} & \cite{DBLP:conf/ideal/BurbidgeRK07} \\\cline{2-5}
& \multirow{3}{*}{Semi-supervised+Active learning} & \multirow{3}{*}{classification} & \multirow{1}{*}{text} & \cite{McCallum:1998:EEP:645527.757765,Tomanek:2009:SAL:1690219.1690291} \\\cline{4-5}
& & & \multirow{1}{*}{image} & \cite{Zhou:2004:EUD:3108498.3108548} \\\cline{4-5}
& & & \multirow{1}{*}{graph} & \cite{Zhu03combiningactive} \\\cline{2-5}
& \multirow{2}{*}{Crowdsourcing} & classification & all & \cite{alfredo:crowdsource,Chang:2017:RCC:3025453.3026044,Mozafari:2014:SUC:2735471.2735474,DBLP:conf/chi/KuleszaACFC14,DBLP:journals/pvldb/WangKFF12,DBLP:conf/cidr/MarcusWMM11,Sheng:2008:GLI:1401890.1401965,DBLP:conf/sigmod/ParameswaranGPPRW12,DBLP:conf/nips/KargerOS11,Dekel2009VoxPC} \\\cline{3-5}
& & regression & all & \cite{DBLP:journals/pvldb/0002KMMO12} \\
\hline
\multirow{2}{*}{Weak supervision} & Data programming & \multirow{1}{*}{classification} &  \multirow{1}{*}{all} & \cite{Zhang15deepdive:a,DBLP:conf/sigmod/Ehrenberg0RFR16,DBLP:conf/nips/RatnerSWSR16,Ratner:2017:SRT:3173074.3173077,DBLP:conf/sigmod/BachRLLSXSRHAKR19,DBLP:conf/sigmod/BringerISRR19,DBLP:conf/icml/BachHRR17,Ratner:2017:SRT:3173074.3173077,Ratner:2018:SMW:3209889.3209898}  \\\cline{2-5}
& Fact extraction & classification & \multirow{1}{*}{text} & \cite{Bollacker:2008:FCC:1376616.1376746,Suchanek:2007:YCS:1242572.1242667,DBLP:conf/cidr/MahdisoltaniBS15,Mausam:2012:OLL:2390948.2391009,Sawadsky:2013:RRC:2486788.2486895,DBLP:conf/emnlp/YahyaWGH14,Dong:2014:KVW:2623330.2623623,Gupta:2014:BOS:2732286.2732288,DBLP:conf/vldb/MeccaCM01,DBLP:conf/www/EtzioniCDKPSSWY04,mitchell2015,carlson-aaai} \\
\hline
\end{tabular}
\caption{A classification of data labeling techniques. Some of the techniques can be used for the same application. For example, for classification on graph data, both self-labeled techniques and label propagation can be used.}
\label{tbl:data_labeling}
\end{table*}

Once enough data has been acquired, the next step is to label individual examples. For instance, given an image dataset of industrial components in a smart factory application, workers can start annotating if there are any defects in the components. In many cases, data acquisition is done along with data labeling. When extracting facts from the Web and constructing a knowledge base, then each fact is assumed to be correct and thus implicitly labeled as true. When discussing the data labeling literature, it is easier to separate it from data acquisition as the techniques can be quite different.

We believe the following categories provide a reasonable view of understanding the data labeling landscape:

\begin{itemize}
   \item {\em Use existing labels: } An early idea of data labeling is to exploit any labels that already exist. There is an extensive literature on semi-supervised learning where the idea is to learn from the labels to predict the rest of the labels.
   \item {\em Crowd-based:} The next set of techniques are based on crowdsourcing. A simple approach is to label individual examples. A more advanced technique is to use active learning where questions to ask are more carefully selected. More recently, many crowdsourcing techniques have been proposed to help workers become more effective in labeling.
   \item {\em Weak labels: } While it is desirable to generate correct labels all the time, this process may be too expensive. An alternative approach is to newly generate less than perfect labels (i.e., weak labels), but in large quantities to compensate for the lower quality. Recently, the latter approach is gaining more popularity as labeled data is scarce in many new applications.
\end{itemize}

Table~\ref{tbl:data_labeling} shows where different labeling approaches fit into the categories. In addition, each labeling approach can be further categorized as follows:

\begin{itemize}
    \item {\em Machine learning task:} In supervised learning, the two categories are classification (e.g., determining whether a piece of text has a positive sentiment) and regression (e.g., estimating the salary of a person). Most of the data labeling research has been focused on classification problems rather than regression problems, possibly because data labeling is simpler in a classification setting. 
    \item {\em Data type:} Depending on the data type (e.g., text, images, and graphs) the data labeling techniques differ significantly. For example, fact extraction from text is very different from object detection on images.
\end{itemize}

\subsection{Utilizing existing labels}

A common setting in machine learning is to have a small amount of labeled data, which is expensive to produce with humans, along with a much larger amount of unlabeled data. Semi-supervised learning techniques~\cite{Zhu2008} exploit both labeled and unlabeled data to make predictions. In a transductive learning setting, the entire unlabeled data is available while in an inductive learning setting, some unlabeled data is available, but the predictions must be on unseen data. Semi-supervised learning is a broad topic, and we focus on a smaller branch of research called {\em self-labeled techniques}~\cite{DBLP:journals/kais/TrigueroGH15} where the goal is to generate more labels by trusting one's own predictions. Since the details are in the survey, we only provide a summary here. In addition to the general techniques, there are graph-based label propagation techniques that are specialized for graph data.

\subsubsection{Classification}

\begin{figure}[t]
\center
  \includegraphics[width=0.4\textwidth]{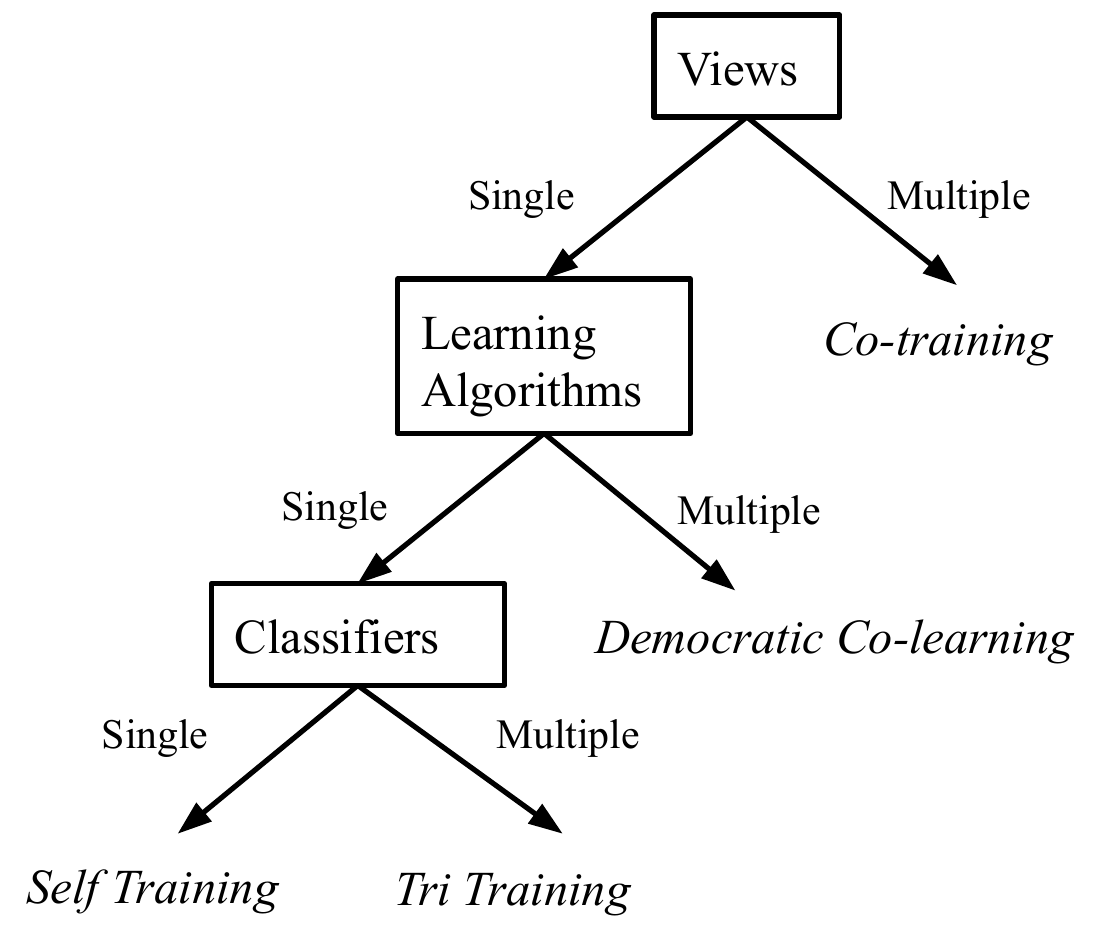}
  \caption{A simplified classification of semi-supervised learning techniques for self labeling according to a survey~\cite{DBLP:journals/kais/TrigueroGH15} using the best-performing techniques regardless of inductive or transductive learning.}
  \label{fig:sslclassification}
\end{figure}

For semi-supervised learning techniques for classification, the goal is to train a model that returns one of multiple possible classes for each example using labeled and unlabeled datasets. We consider the best-performing techniques in a  survey that focuses on labeling data~\cite{DBLP:journals/kais/TrigueroGH15}, which are summarized in Figure~\ref{fig:sslclassification}. The performance results are similar regardless of using transductive or inductive learning.

The simplest class of semi-supervised learning techniques train one model using one learning algorithm on one set of features. For example, Self-training~\cite{Yarowsky:1995:UWS:981658.981684} initially trains a model on the labeled examples. The model is then applied to all the unlabeled data where the examples are ranked by the confidences in their predictions. The most confident predictions are then added into the labeled examples. This process repeats until all the unlabeled examples are labeled.

The next class trains multiple classifiers by sampling the training data several times and training a model for each sample. For example, Tri-training~\cite{Zhou:2005:TEU:1092713.1092809} initially trains three models on the labeled examples using Bagging for the ensemble learning algorithm. Then each model is updated iteratively where the other two models make predictions on the unlabeled examples, and only the examples with the same predictions are used in conjunction with the original labeled examples to re-train the model. The iteration stops when no model changes. Finally, the unlabeled examples are labeled using majority voting where at least two models must agree with each other.

The next class uses multiple learning algorithms. For example, Democratic Co-learning~\cite{DBLP:conf/ictai/ZhouG04} uses a set of different learning algorithms (in the experiments, they are Naive Bayes, C4.5, and 3-nearest neighbor) are used to train a set of classifiers separately on the same training data. Predictions on new examples are generated by combining the results of the three classifiers using weighted voting. The new labels are then added to the training set of the classifiers whose predictions are different from the majority results. This process repeats until no more data is added to the training data of a classifier.

The final class uses multiple views, which are subsets of features that are conditionally independent given the class. For example, Co-training~\cite{Blum:1998:CLU:279943.279962} splits the feature set into two sufficient and redundant views, which means that one set of features is sufficient for learning and independent of learning with the other set of features given the label. For each feature set, a model is trained and then used to teach the model trained on the other feature set. The co-trained models can minimize errors by maximizing their agreements over the unlabeled examples.

According to the survey, these algorithms result in similar transductive or inductive accuracies when averaged on 55 datasets from the UCI~\cite{Dua:2017} and KEEL dataset~\cite{journals/mvl/Alcala-FdezFLDG11} repositories. Here accuracy is defined as the portion of classifications by the trained model that are correct. However, not all of these techniques are generally-applicable where one can plug in any machine learning algorithm and use any set of features. First, Co-training assumes that the feature can be divided into two subsets that are conditionally independent given the label (also called sufficient and redundant views), which is not always possible. Second, Democratic Co-learning assumes three different machine learning algorithms, which is not always possible if there is only one algorithm to use. 

\subsubsection{Regression}

Relatively less research has been done for semi-supervised learning for regression where the goal is to train a model that predicts a real number given an example. Co-regularized least squares regression~\cite{Brefeld:2006:ECL:1143844.1143862} is a least squares regression algorithm based on the co-learning approach. Another co-regularized framework~\cite{Sindhwani05aco-regularized} utilizes sufficient and redundant views similar to Co-training. Co-training regressors~\cite{Zhou:2005:SRC:1642293.1642439} uses two $k$-nearest neighbor regressors with different distance metrics. In each iteration, a regressor labels the unlabeled data that can be labeled most confidently by the other regressor. After the iterations, the final prediction of an example is made by averaging the regression estimates by the two regressors. Co-training Regressors can be extended by using any other base regressor.

\subsubsection{Graph-based Label Propagation}
Graph-based label propagation techniques also start with limited sets of labeled examples, but exploit the graph structure of examples based on their similarities to infer the labels of the remaining examples. For example, if an image is labeled as a dog, then similar images down the graph can also be labeled as dogs with some probability. The further the distance, the lower the probability of label propagation. Graph-based label propagation has applications in computer vision, information retrieval, social networks, and natural language processing. Zhu et al.~\cite{Zhu:2003:SLU:3041838.3041953} proposed a semi-supervised learning based on a Gaussian random field model where the unlabeled and labeled examples form a weighted graph. The mean of the field is characterized in terms of harmonic functions and can be efficiently computed using matrix methods or belief propagation. The MAD-Sketch algorithm~\cite{DBLP:conf/aistats/TalukdarC14} was proposed to further reduce the space and time complexities of graph-based SSL algorithms using count-min sketching. In particular, the space complexity per node is reduced from $O(m)$ to $O(\text{log}\ m)$ under certain conditions where $m$ is the number of distinct labels, and a similar improvement is achieved for the time complexity. Recently, a family of algorithms called EXPANDER~\cite{DBLP:conf/aistats/RaviD16} were proposed to further reduce the space complexity per node to $O(1)$ and compute the MAD-Sketch algorithm in a distributed fashion.

\subsection{Crowd-based techniques}
\label{sec:crowdbasedtechniques}

The most accurate way to label examples is to do it manually. A well known use case is the ImageNet image classification dataset~\cite{DBLP:conf/cvpr/DengDSLL009} where tens of millions of images were organized according to a semantic hierarchy by WordNet using Amazon Mechanical Turk. However, ImageNet is an ambitious project that took years to complete, which most machine learning users cannot afford for their own applications.  Traditionally, active learning has been a key technique in the machine learning community for carefully choosing the right examples to label and thus minimize cost. More recently, crowdsourcing techniques for labeling have been proposed where there can be many workers who are not necessarily experts in labeling. Hence, there is more emphasis on how to assign tasks to workers, what interfaces to use, and how to ensure high quality labels. Recent commercial tools vary in what services they provide for labeling. For example, Amazon SageMaker~\cite{sagemaker} supports labeling based on active learning, Google Cloud AutoML~\cite{googlecloudautoml} provides a manual labeling service, and Microsoft Custom Vision~\cite{microsoftcustomvision} requires labels from the user. While crowdsourcing data labeling is closely related to crowdsourcing data acquisition, the individual techniques are different.

\subsubsection{Active Learning}
\label{sec:activelearning}

Active learning focuses on selecting the most ``interesting'' unlabeled examples to give to the crowd for labeling. The workers are expected to be very accurate, so there is less emphasis on how to interact with those with less expertise. While some references view active learning as a special case of semi-supervised learning, the key difference is that there is a human-in-the-loop. The key challenge is choosing the right examples to ask given a limited budget. One downside of active learning is that the examples are biased to the training algorithm and cannot be reused. Active learning is covered extensively in other surveys~\cite{DBLP:series/synthesis/2012Settles,DBLP:reference/sp/2015rsh}, and we only cover the most prominent techniques here.
\customparagraph{Uncertain Examples } Uncertainty Sampling~\cite{Lewis:1994:SAT:188490.188495} is the simplest in active learning and chooses the next unlabeled example that the model prediction is most uncertain. For example, if the model is a binary classifier, uncertainty sampling chooses the example whose probability is nearest to 0.5. If there are more than three class labels, we could choose the example whose prediction is the least confident. The downside of this approach is that it throws away the information of all the other possible labels. So an improved version called margin sampling is to choose the example whose probability difference between the most and second-most probable labels is the largest. This method can be further generalized using entropy as the uncertainty measure where entropy is an information-theoretic measure for the amount of information to encode a distribution. 



Query-by-Committee~\cite{Seung:1992:QC:130385.130417} extends uncertainty sampling by training a committee of models on the same labeled data. Each model can vote when labeling each example, and the most informative example is considered to be the one where the most models disagree with each other. More formally, this approach minimizes the {\em version space}, which is the space of all possible classifiers that give the same classification results as (and are thus consistent with) the labeled data. The challenge is to train models that represent different regions of the version space and have some amount of disagreement. Various methods have been proposed~\cite{DBLP:series/synthesis/2012Settles}, but there does not seem to be a clear winner. One general method is called query-by-bagging~\cite{Abe:1998:QLS:645527.657478}, which uses bagging as an ensemble learning algorithm and trains models on bootstrap samples. There is no general agreement on the best number of models to train, which is application-specific.

Both uncertainty sampling and query-by-committee focus on individual examples instead of the entire set of examples, they run into the danger of choosing examples that are outliers according to the example distribution. Density weighting~\cite{Settles:2008:AAL:1613715.1613855} is a way to improve the above techniques by choosing instances that are not only uncertain or disagreeing, but also representative of the example distribution.
\customparagraph{Decision Theoretic Approaches } Another line of active learning performs decision-theoretic approaches. Decision theory is a  framework for making decision under uncertainty using states and actions to optimize some objective function. In the context of active learning, the objective could be to choose an example that maximizes the estimated model accuracy~\cite{Settles:2007:MAL:2981562.2981724}. Another possible objective is reducing generalization  error~\cite{Roy:2001:TOA:645530.655646}, which is estimated as follows: if the measure is log loss, then the entropy of the predicted class distribution is considered the error rate; if the measure is 0-1 loss, the maximum probability among all classes is the error rate. Each example to label is chosen by taking a sample of the unlabeled data and choosing the example that minimizes the estimated error rate.
\customparagraph{Regression } Active learning techniques can also be extended to regression problems. For uncertainty sampling, instead of computing the entropy of classes, one can compute the output variance of the predictions and select the examples with the highest variance. Query-by-committee can also be extended to regression~\cite{DBLP:conf/ideal/BurbidgeRK07} by training a committee of models and selecting the examples where the variance among the committee's predictions is the largest. This approach is said to work well when the bias of the models is small. Also, this approach is said to be robust to overspecification, which reduces the chance of overfitting.
\customparagraph{Self and Active Learning Combined } The data labeling techniques we consider are complementary to each other and can be used together. In fact, semi-supervised learning and active learning have a history of being used together~\cite{McCallum:1998:EEP:645527.757765,10.1007/978-3-319-12637-1_27,Zhou:2004:EUD:3108498.3108548,Zhu03combiningactive,Tomanek:2009:SAL:1690219.1690291}. A key observation is that the two techniques solve opposite problems where semi-supervised learning finds the predictions with the highest confidence and adds them to the labeled examples while active learning finds the predictions with the lowest confidence (using uncertainty sampling, query-by-committee, or density-weighted method) and sends them for manual labeling.

There are various ways semi-supervised learning can be used with active learning. McCallum and Nigam~\cite{McCallum:1998:EEP:645527.757765} improves the Query-By-Committee (QBC) technique and combines it with Expectation-Maximization (EM), which effectively performs semi-supervised learning. Given a set of documents for training data, active learning is done by selecting the documents that are closer to others (and thus representative), but have committee disagreement, for labeling. In addition, the EM algorithm is used to further infer the rest of the labels. The active learning and EM can either be done separately or interleaved. Tomanek and Hahn~\cite{Tomanek:2009:SAL:1690219.1690291} propose semi-supervised active learning (SeSAL) for sequence labeling tasks, which include POS tagging, chunking, or named entity recognition (NER). Here the examples are sequences of text. The idea is to use active learning for the subsequences that have the highest training utility within the selected sentences and use semi-supervised learning to automatically label the rest of the subsequences. The utility of a subsequence is highest when the current model is least confident about the labeling.

Zhou et al.~\cite{Zhou:2004:EUD:3108498.3108548} proposes the semi-supervised active image retrieval (SSAIR) approach where the focus is on image retrieval. SSAIR is inspired by the co-training method where initially two classifiers are trained from the labeled data. Then each learner passes the most relevant/irrelevant images to the other classifier. The classifiers are then retrained with the additional labels, and their results are combined. The images that still have low confidence are selected to be labeled by humans. 

Zhu et al.~\cite{Zhu03combiningactive} combines semi-supervised and active learning under a Gaussian random field model. The labeled and unlabeled examples are represented as vertices in a graph where edges are weighted by similarities between examples. This framework enables one to compute the next question that minimizes the expected generalization error efficiently for active learning. Once the new labels are added to the labeled data, semi-supervised learning is performed using harmonic functions. 


\subsubsection{Crowdsourcing}
\label{sec:labeling-crowdsourcing}

In comparison to active learning, the crowdsourcing techniques here are more focused on running tasks with many workers who are not necessarily labeling experts. As a result, workers may make mistakes, and there is a heavy literature~\cite{DBS-044,6488672,Daniel:2018:QCC:3177787.3148148,7420720,Crescenzi:2015:CLS:2735847.2735894,Chang:2017:RCC:3025453.3026044,DBLP:journals/pacmhci/SchaekermannGLL18} on improving the interaction with workers, evaluating workers so they are reliable, reducing any bias that the workers may have, and aggregating the labeling results while resolving any ambiguities among them.
\customparagraph{User Interaction } A major challenge in user interaction is to effectively provide instructions to workers on how to perform the labeling. The traditional approach is to provide some guidelines for labeling to the workers up front and then let them make a best effort to follow them. However, the guidelines are often incomplete and do not cover all possible scenarios, leaving the workers in the dark. Revolt~\cite{Chang:2017:RCC:3025453.3026044} is a system that attempts to fix this problem through collaborative crowdsourcing. Here workers work in three steps: {\em Voting} where workers vote just like in traditional labeling, {\em Explaining} where workers justify their rational for labeling, and {\em Categorize} where workers review explanations from other workers and tag any conflicting labels. This information can then be used to make post-hoc judgements of the label decision boundaries. Another approach is to provide better tools to assist workers to organize their concepts, which may evolve as more examples are labeled~\cite{DBLP:conf/chi/KuleszaACFC14}.

In addition, providing the right labeling interface is critical for workers to perform well. The challenge is that each application may have a different interface that works best. We will not cover all the possible applications, but instead illustrate a line of research for the problem of entity resolution where the goal is to find records in a database that refer to the same real-world entity. Here the label is whether two (or more) records are the same or not. Just for this problem, there is a line of research on providing the best interface for comparisons. CrowdER~\cite{DBLP:journals/pvldb/WangKFF12} provides two types of interfaces to compare records: pair-based and cluster-based. Qurk~\cite{DBLP:conf/cidr/MarcusWMM11} uses a mapping interface where multiple records on one side are matched with records on the other side. Qurk uses a combination of comparison or rating tasks to accelerate labeling.
\customparagraph{Quality control } Controlling the quality of data labeling by the crowd is important because the workers may vary significantly in their abilities to provide labels. A simple way to ensure quality is to repeatedly label the same example using multiple workers and perhaps take a majority voting at the end. However, there are more sophisticated approaches as well. Get another label~\cite{Sheng:2008:GLI:1401890.1401965} and Crowdscreen~\cite{DBLP:conf/sigmod/ParameswaranGPPRW12} actively solicit labels while Karger et al.~\cite{DBLP:conf/nips/KargerOS11} passively collects data and runs the expectation maximization algorithm. Vox Populi~\cite{Dekel2009VoxPC} proposes techniques for pruning low-quality workers that can achieve better labeling quality without having to repeatedly label examples. 
\customparagraph{Scalability } Scaling up crowdsourced labeling is another important challenge. While traditional active learning techniques were proposed for this purpose, more recently the data management community has started to apply systems techniques for further scaling the algorithms to large datasets. In particular, Mozafari et al.~\cite{Mozafari:2014:SUC:2735471.2735474} proposes active learning algorithms that can run in parallel. One algorithm (called {\em Uncertainty}) selects examples that the current classifier is most uncertain about. A more sophisticated algorithm (called {\em MinExpError}) combines the current model's accuracy with the uncertainty. A key idea is the use of bootstrap theory, which makes the algorithms applicable to any classifier and also enables embarassingly-parallel processing.
\customparagraph{Regression } In comparison to crowdsourcing research for classification tasks, less attention has been given to regression tasks. Marcus et al.~\cite{DBLP:journals/pvldb/0002KMMO12} solves the problem of selectivity estimation in a crowdsourced database. The goal is to estimate the fraction of records that satisfy some property by asking workers questions.


\subsection{Weak supervision}

As machine learning is used in a wider range of applications, it is mostly the case that there is not enough labeled data. For example, in a smart factory setting, any new product will have no labels for training a model for quality control. As a result, weak supervision techniques~\cite{weaksupervision,DBLP:conf/cidr/RatnerHR19,weaklysupervisedlearning} have become increasingly popular where the idea is to semi-automatically generate large quantities of labels that are not as accurate as manual labels, but good enough for the trained model to obtain a reasonably-high accuracy. This approach is especially useful when there are large amounts of data, and manual labeling becomes infeasible. In the next sections, we discuss the recently proposed data programming paradigm and fact extraction techniques.

\subsubsection{Data Programming}

\begin{figure}[t]
\center
  \includegraphics[width=0.5\textwidth]{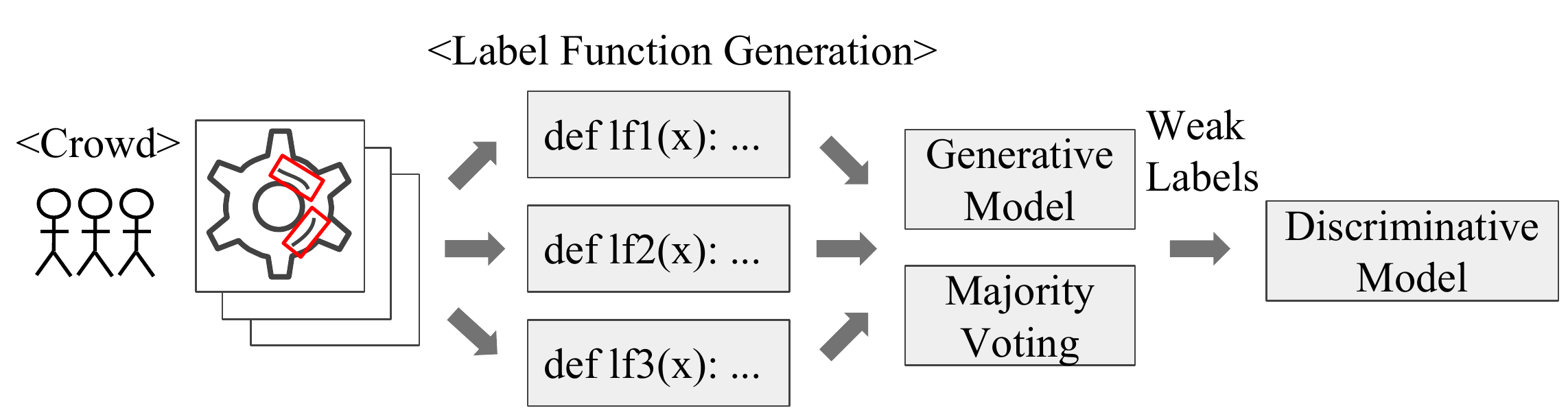}
  \caption{A workflow of using data programming for a smart factory application. In this scenario, Sally is using crowdsourcing to annotate defects on component images. Next, the annotations can be automatically converted to labeling functions. Then the labeling functions are combined either into a generative model or using majority voting. Finally, the combined model generates weak labels that are used to train a discriminative model.}
  \label{fig:dataprogramming}
\end{figure}

As data labeling at scale becomes more important especially for deep learning applications, data programming~\cite{DBLP:conf/nips/RatnerSWSR16} has been proposed as a solution for generating large amounts of weak labels using multiple labeling functions instead of individual labeling. Figure~\ref{fig:dataprogramming} illustrates how data programming can be used for Sally's smart factory application. A labeling function can be any computer program that either generates a label for an example or refrains to do so. For example, a labeling function that checks if a tweet has a positive sentiment may check if certain positive words appear in the text. Since a single labeling function by itself may not be accurate enough or not be able to generate labels for all examples, multiple labeling functions are implemented and combined into a generative model, which is then used to generate large amounts of weak labels with reasonable quality. Alternatively, voting methods like majority voting can be used to combine the labeling functions. Finally, a noise-aware discriminative model is trained on the weak labels. Data programming has been implemented in the state-of-the-art Snorkel system~\cite{Ratner:2017:SRT:3173074.3173077}, which is becoming increasingly popular in the industry~\cite{DBLP:conf/sigmod/BachRLLSXSRHAKR19,DBLP:conf/sigmod/BringerISRR19}.

 Data programming has advantages both in terms of model accuracy and usability. A key observation for generating weak labels is that training a discriminative model on large amounts of weak labels may result in higher accuracy than training with fewer manual labels. In terms of usability, implementing labeling functions can be an intuitive process for humans compared to feature engineering in traditional machine learning~\cite{Ehrenberg:2016:DPD:2939502.2939515}.

What makes data programming effective is the way it combines multiple labeling functions into the generative model by fitting a probabilistic graphical model. A na\"ive approach to combine labeling functions is to take a majority vote. However, this approach cannot handle pathological cases where many labeling functions are near identical, which defeats the purpose of majority voting. Instead, labeling functions that are more correlated with each other will have less influence on the predicted label. In addition, labeling functions that are outliers are also trained to have less influence in order to cancel out the noise. Theoretical analysis~\cite{DBLP:conf/icml/BachHRR17} shows that, if the labeling functions are reasonably accurate, then the predictions made by the generative model becomes arbitrarily close to the true labels.

Several systems have been developed for data programming. DeepDive~\cite{Zhang15deepdive:a} is a precursor of data programming and supports fast knowledge base construction on dark data using information extraction. DeepDive effectively uses humans to extract features, implement supervision rules using a declarative language (similar to labeling functions), and supports incremental maintenance of inference. DDLite~\cite{DBLP:conf/sigmod/Ehrenberg0RFR16} is the first system to use data programming and supports rapid prototyping of labeling functions through an interactive interface. The primary application is information extraction, and DDLite has been used to extract chemicals, diseases, and anatomical named entities. Compared to DeepDive, it has a simpler Python syntax and does not require complex setup involving databases. Snorkel~\cite{Ratner:2017:SFT:3035918.3056442,Ratner:2017:SRT:3173074.3173077} is the most recent system for data programming. Compared to DDLite, Snorkel is a full-fledged product that is widely used in the industry. Snorkel enables users to use weak labels from all available weak label sources, supports any type of classifier, and provides rapid results in response to the user's input. More recently, Snorkel has been extended to solve massively multi-task learning~\cite{Ratner:2018:SMW:3209889.3209898}.

Data programming is designed for classification and does not readily support regression. To make an extension to regression, the labeling functions must return real numbers instead of discrete values. In addition, the probabilistic graphical model used for training generative models must be able to return a continuous distribution of possible values for labels.

\subsubsection{Fact Extraction}

Another way to generate weak labels is use fact extraction. Knowledge bases contain facts that are extracted from various sources including the Web. A fact could describe an attribute of an entity (e.g., $\langle$Germany, {\it capital}, Berlin$\rangle$). The facts can be considered positively-labeled examples, which can be used as seed labels for distant supervision~\cite{Mintz:2009:DSR:1690219.1690287} when generating weak labels. It is worth mentioning that fact extraction roots from the broader topic of information extraction where the goal is to extract structured data from the Web~\cite{DBLP:journals/kbs/FerraraMFB14}. Early work include RoadRunner~\cite{DBLP:conf/vldb/MeccaCM01}, which compares HTML pages to generates wrappers and KnowItAll~\cite{DBLP:conf/www/EtzioniCDKPSSWY04}, which uses extraction rules and a search engine to identify and rank facts. Since then, the subsequent works have become more sophisticated and also attempt to organize the extract facts into knowledge bases.

We now focus on fact extraction techniques for knowledge bases. If precision is critically important, then manual curation should be part of the knowledge base construction as in Freebase~\cite{Bollacker:2008:FCC:1376616.1376746} and Google Knowledge Graph. Otherwise, the extraction techniques depend on the data source. YAGO~\cite{Suchanek:2007:YCS:1242572.1242667,DBLP:conf/cidr/MahdisoltaniBS15} extracts facts from Wikipedia using classes in WordNet. Ollie~\cite{Mausam:2012:OLL:2390948.2391009}, ReVerb~\cite{Sawadsky:2013:RRC:2486788.2486895}, and ReNoun~\cite{DBLP:conf/emnlp/YahyaWGH14} and open information extraction systems that apply patterns to Web text. Knowledge Vault~\cite{Dong:2014:KVW:2623330.2623623} also extracts from Web content, but combines facts from text, tabular data, page structure, and human annotations. Biperpedia~\cite{Gupta:2014:BOS:2732286.2732288} extracts the attributes of entities from a query stream and Web text.

The Never-Ending Language Learner (NELL) system~\cite{mitchell2015,carlson-aaai} continuously extracts structured information from the unstructured Web and constructs a knowledge base that consists of entities and relationships. Initially, NELL starts with a seed ontology, which contains entities of classes (e.g., person, fruit, and emotion) and relationships among the entities (e.g., $playsOnTeam(athlete, sportsTeam)$ and $playsInstrument(musician, instrument)$). NELL analyzes hundreds of millions of Web pages and identifies new entities in the given classes as well as entity pairs of the relationships by matching patterns on their surrounding phrases. The resulting entities and relationships can then be used as the next training data for constructing even more patterns. NELL has been collecting facts continuously since 2010. The extraction techniques can be viewed as distant supervision generating weak labels.

\section{Using Existing Data and Models}
\label{sec:without-data}

An alternative approach to acquiring new data and labeling it is to improve the labeling of any existing datasets or improving the model training. This approach makes sense for a number of scenarios. First, it may be difficult to find new datasets because the application is too novel or non-trivial for others to have produced datasets. Second, simply adding more data may not significantly improve the model's accuracy anymore. In this case, re-labeling or cleaning the existing data may be the faster way to increase the accuracy. Alternatively, the model training can be made more robust to noise and bias, or the model can be trained from an existing model using transfer learning techniques. In the following sections, we explore techniques for improving existing labels and improving existing models. The techniques are summarized in Table~\ref{tbl:improving_existing}.

\begin{table}[t]
\renewcommand{\arraystretch}{1.3}
\centering
\begin{tabular}{cc}
\hline
{\bf Task}  & {\bf Techniques} \\
\hline \hline
\multirow{2}{*}{Improve Data} & Data Cleaning~\cite{DBLP:journals/pvldb/RekatsinasCIR17,DBLP:journals/pvldb/KrishnanWWFG16,DBLP:journals/corr/abs-1711-01299,dolatshah2018,DBLP:conf/vldb/RamanH01,DBLP:conf/chi/KandelPHH11,DBLP:conf/avi/KandelPPHH12,DBLP:conf/pldi/HarrisG11,DBLP:conf/sigmod/Tae19data} \\\cline{2-2}
& Re-labeling~\cite{Sheng:2008:GLI:1401890.1401965} \\
\hline
\multirow{2}{*}{Improve Model} & Robust Against Noise~\cite{conf/cvpr/XiaoXYHW15,DBLP:conf/iccv/ChenG15,He:2009:LID:1591901.1592322,Chawla:2002:SSM:1622407.1622416,DBLP:journals/corr/GoodfellowSS14} \\\cline{2-2}
 & Transfer Learning~\cite{Pan:2010:STL:1850483.1850545,DBLP:journals/jbd/WeissK016,tfhub,DBLP:journals/pami/Fei-FeiFP06,DBLP:conf/naacl/RuderPSW19,DBLP:conf/nips/YosinskiCBL14,DBLP:journals/jbd/DayK17}   \\\hline
\end{tabular}
\caption{A classification of techniques for improving existing data and models.}
\label{tbl:improving_existing}
\end{table}

\subsection{Improving Existing Data}

A major problem in machine learning is that the data can be noisy and the labels incorrect. This problem occurs frequently in practice, so production machine learning platforms like TensorFlow Extended (TFX)~\cite{DBLP:conf/kdd/BaylorBCFFHHIJK17} have separate components~\cite{tfdv} to reduce data errors as much as possible though analysis and validation. In case the labels are also noisy, re-labeling the examples becomes necessary as well. We explore recent advances in data cleaning with a focus on machine learning and then techniques for re-labeling. 

\subsubsection{Data Cleaning}

It is common for the data itself to be noisy. For example, some values may be out of range (e.g., a latitude value is beyond [-90, 90]) or use different units by mistake (e.g., some intervals are in hours while other are in minutes). There is a heavy literature on various integrity constraints (e.g., domain constraints, referential integrity constraints, and functional dependencies) that can improve data quality as well. HoloClean~\cite{DBLP:journals/pvldb/RekatsinasCIR17} is a state-of-art data cleaning system that uses quality rules, value correlations, and reference data to build a probabilistic model that captures how the data was generated. HoloClean then generates a probabilistic program for repairing the data. In addition, various interactive data cleaning tools~\cite{DBLP:conf/vldb/RamanH01,DBLP:conf/chi/KandelPHH11,DBLP:conf/avi/KandelPPHH12,DBLP:conf/pldi/HarrisG11} have been proposed to convert data into a better form for machine learning.

An interesting line of recent work is cleaning techniques with the explicit intention of improving machine learning results. ActiveClean~\cite{DBLP:journals/pvldb/KrishnanWWFG16} is a model training framework that iteratively suggests samples of data to clean based on how much the cleaning improves the model accuracy and the likelihood that the data is dirty. An analyst can then perform transformations and filtering to clean each sample. ActiveClean treats the training and cleaning as a form of stochastic gradiant descent and uses convex-loss models (SVMs, linear and logistic regression) to guarantee global solutions for clean models. BoostClean~\cite{DBLP:journals/corr/abs-1711-01299} solves an important class of inconsistencies where an attribute value is outside an allowed domain. BoostClean takes as input a dataset and a set of functions that can detect these errors and repair functions that can fix them. Each pair of detection and repair functions can produce a new model trained on the cleaned data. BoostClean uses statistical boosting to find the best ensemble of pairs that maximize the final model's accuracy. Recently, {\sc TARS}~\cite{dolatshah2018} was proposed to solve the problem of cleaning crowdsourced labels using oracles. {\sc TARS} provides two pieces of advice. First, given test data with noisy labels, it uses an estimation technique to predict how well the model may perform on the true labels. The estimation is shown to be unbiased, and confidence intervals are computed to bound the error. Second, given training data with noisy labels, {\sc TARS} determines which examples to send to an oracle in order to maximize the expected model improvement of cleaning each noisy label. More recently, {\sc MLClean}~\cite{DBLP:conf/sigmod/Tae19data} has been proposed to integrate three data operations: traditional data cleaning, model unfairness mitigation where the goal is to remove data bias that causes model fairness, and data sanitization where the goal is to remove data poisoning.

\subsubsection{Re-labeling}

Trained models are only as good as their training data, and it is important to obtain high quality data labels. Simply labeling more data may not improve the model accuracy further. Indeed, Sheng et al.~\cite{Sheng:2008:GLI:1401890.1401965} shows that, if the labels are noisy, then the model accuracy plateaus from some point and does not increase further, no matter how many more labeling is done. The solution is to improve the quality of existing labels. The authors show that repeated labeling using workers of certain individual qualities can significantly improve model accuracy where a straightforward round robin approach already give substantial improvements, and being more selective in labeling gives even better results.

\subsection{Improving Models}

In addition to improving the data, there are also ways to improve the model training itself. Making the model training more robust against noise or bias is an active area of research. Another popular approach is to use transfer learning where previously-trained models are used as a starting point to train the current model.

\subsubsection{Robust Against Noise and Bias}

A common scenario in machine learning is that there is a large number of noisy or even adversarial labels and a relatively smaller number of clean labels. Simply discarding the noisy labels will result in reduced training data, which is not desirable for complex models. Hence, there has been extensive research (see the survey~\cite{DBLP:journals/tnn/FrenayV14}) on how to make the model training still use noisy labels by becoming more robust. For specific techniques, Xiao et al.~\cite{conf/cvpr/XiaoXYHW15} propose a general framework for training convolutional neural networks on images with a small number of clean labels and many noisy labels. The idea is to model the relationships between images, class labels, and label noises with a probabilistic graphical model and integrate it into the model training. Label noise is categorized into two types: confusing noise, which is caused by confusing content in the images, and pure random noise, which is caused by technical bugs like mismatches between images and their surrounding text. The true labels and noise types are treated as latent variables, and an EM algorithm is used for inference. Webly supervised learning~\cite{DBLP:conf/iccv/ChenG15} is a technique for training a convolutional neural network on clean and noisy images on the Web. First, the model is trained on top-ranked images from search engines, which tend to be clean because they are highly-ranked, but also biased in the sense that objects tends to be centered in the image with a clean background. Then relationships are discovered among the clean images, which are then used to adapt the model to more noisier images that are harder to classify. This method suggests that it is worth training on easy and hard data separately.



Goodfellow et al.~\cite{DBLP:journals/corr/GoodfellowSS14} take a different approach where they explain why machine learning models including neural networks may misclassify adversarial examples. While previous research attempts to explain this phenomenon by focusing on nonlinearity and overfitting, the authors show that it is the model's linear behavior in high-dimensional spaces that makes it vulnerable. That is, making many small changes on the features of an example can result in a large change to the output prediction. As a result, generating large amounts of adversarial examples becomes easier using linear perturbation.

Even if the labels themselves are clean, it may be the case that the labels are imbalanced. $SMOTE$~\cite{Chawla:2002:SSM:1622407.1622416} performs over-sampling for minority classes that need more examples. Simply replicating examples may lead to overfitting, so the over-sampling is done by generating synthetic examples using the minority examples and their nearest neighbors. He and Garcia~\cite{He:2009:LID:1591901.1592322} provide a comprehensive survey on learning from imbalanced data.


\subsubsection{Transfer Learning}

Transfer learning is a popular approach for training models when there is not enough training data or time to train from scratch. A common technique is to start from an existing model that is well trained (also called a {\em source task}), one can incrementally train a new model (a {\em target task}) that already performs well. For example, a convolutional neural networks like AlexNet~\cite{DBLP:conf/nips/KrizhevskySH12} and VGGNet~\cite{DBLP:journals/corr/SimonyanZ14a} can be used to train a model for a different, but related vision problem. Recently, Google announced TensorFlow Hub~\cite{tfhub}, which enables users to easily re-use an existing model to train an accurate model, even with a small dataset. Also, Google Cloud AutoML~\cite{googlecloudautoml} provides transfer learning as a service. From a data management perspective, an interesting question is how these existing tools can be extended to index the metadata of models and provide search as a service, just like for datasets. The metadata for models may be quite different than metadata for data because one needs to determine if a model can be used for transfer learning in her own application. In addition to using pre-trained models, another popular technique mainly used in Computer Vision is few-shot learning~\cite{DBLP:journals/pami/Fei-FeiFP06} where the goal is to extend existing models to handle new classes using zero or more examples. Since transfer learning is primarily a machine learning topic that does not significantly involve data management, we only summarize the high-level ideas based on surveys~\cite{Pan:2010:STL:1850483.1850545,DBLP:journals/jbd/WeissK016}. There are studies of transfer learning techniques in the context of NLP~\cite{DBLP:conf/naacl/RuderPSW19}, Computer Vision~\cite{DBLP:conf/nips/YosinskiCBL14}, and deep learning~\cite{DBLP:conf/icann/TanSKZYL18} as well.

An early survey of transfer learning~\cite{Pan:2010:STL:1850483.1850545} identifies three main research issues in transfer learning: what to transfer, how to transfer, and when to transfer. That is, we need to decide what part of knowledge can be transferred, what methods should be used to transfer the knowledge, and whether transferring this knowledge is appropriate and does not have any negative effect. {\em Inductive transfer learning} is used when the source task and target task are different while the two domains may or may not be the same. Here a task can be categorizing a document while a domain could be a set of university webpages to categorize. {\em Transductive transfer learning} is used when the source and target tasks are the same, but the domains are different. {\em Unsupervised transfer learning} is similar to inductive transfer learning where the source and target tasks are different, but uses unsupervised learning tasks like clustering and dimensionality reduction.

The three approaches above can also be divided based on what to transfer. {\em Instance-based transfer learning} assumes that the examples of the source can be re-used in the target by re-weighting them. {\em Feature-representation transfer learning} assumes that the features that represent the data of the source task can be used to represent the data of the target task. {\em Parameter transfer learning} assumes that the source and target tasks share some parameters or prior distributions that can be re-used. {\em Relational knowledge transfer learning} assumes that certain relationships within the data of the source task can be re-used in the target task.

More recent surveys~\cite{DBLP:journals/jbd/WeissK016,DBLP:journals/jbd/DayK17} classify most of the traditional transfer learning techniques as {\em homogeneous transfer learning} where the feature spaces of the source and target tasks are the same. In addition, the surveys identify a relatively new class of techniques called {\em heterogeneous transfer learning} where the feature spaces are different, but the source and target examples are extracted from the same domain. Heterogeneous transfer learning largely falls into two categories: asymmetric and symmetric transformation. In an asymmetric approach, features of the source task are transformed to the features of the target task. In a symmetric approach, the assumption is that there is a common latent feature space that unifies the source and target features. Transfer learning has been successfully used in many applications including text sentiment analysis, image classification, human activity classification, software defect detection, and multi-language text classification.

\section{Putting Everything Together}
\label{sec:which-technique}

We now return to Sally's scenario and provide an end-to-end guideline for data collection (summarized as the workflow in Figure~\ref{fig:workflow}). If there is no or little data to start with then Sally would need to acquire datasets. She can either search for relevant datasets either on the Web or within the company data lake, or decide to generate a dataset herself by installing camera equipment for taking photos of the products within the factory. If the products also had some metadata, Sally could also augment that data with external information about the product.

Once the data is available, then Sally can choose among the labeling techniques using the categories discussed in Section~\ref{sec:data-labeling}. If there are enough existing labels, then self labeling using semi-supervised learning is an attractive option. There are many variants of self labeling depending on the assumptions on the model training as we studied. If there are not enough labels, Sally can decide to generate some using the crowd-based techniques using a budget. If there are only a few experts available for labeling, active learning may be the right choice, assuming that the important examples that influence the model can be narrowed down. If there are many workers who do not necessarily have expertise, general crowdsourcing methods can be used. If Sally does not have enough budget for crowd-based methods or if it is simply not worth the cost, and if the model training can tolerate weak labels, then weak supervision techniques like data programming and label propagation can be used.

If Sally has existing labels, she may also want to make sure whether they can be improved in quality. If the data is noisy or biased, then the various data cleaning techniques can be used. If there are existing models for product quality through tools like TensorFlow Hub~\cite{tfhub}, they can be used to further improve the model using transfer learning.

Through our experience, we also realize that it is not always easy to determine if there is enough data and labels. For example, even if the dataset is small or there are few labels, as long as the distribution of data is easy to learn, then automatic approaches like semi-supervised learning will do the job better than manual approaches like active learning. Another hard-to-measure factor is the amount of human effort needed. When comparing active learning versus data programming, we need to compare the tasks of labeling examples and implementing labeling functions, which are quite different. Depending on the application, implementing a program on examples can range from trivial (e.g., look for certain keywords) to almost impossible (e.g., general object detection). Hence, even if data programming is an attractive option, one must determine the actual effort of programming, which cannot be determined with a few yes or no questions.

Another thing to keep in mind is how the labeling techniques tradeoff accuracy and scalability. Manual labeling is obviously the most accurate, but least scalable. Active learning scales better than the manual approach, but is still limited to how fast humans can label. Data programming produces weak labels, which tend to have lower accuracy than manual labels. On the other hand, data programming can scale better than active learning assuming that the initial cost of implementing labeling functions and debugging them is reasonable. Semi-supervised learning obviously scales the best with automatic labeling. The labeling accuracy depends on the accuracy of the model trained on existing labels. Combining self labeling with active learning is a good example of taking the best of both worlds.


\section{Future Research Challenges}
\label{sec:future-work}

Although data collection was traditionally a topic in the machine learning, as the amount of training data is increasing, data management research is becoming just as relevant, and we are observing a convergence of the two disciplines. As such, there needs to be more awareness on how the research landscape will evolve for both communities and more effort to better integrate the techniques.
\customparagraph{Data Evaluation}
An open question is how to evaluate whether the right data was collected with sufficient quantity. First, it may not be clear if we have found the best datasets for a machine learning task and whether the amount of data is enough to train a model with sufficient accuracy. In some cases, there may be too many datasets, and simply collecting and integrating all of them may have a negative affect on model training. As a result, selecting the right datasets becomes an important problem. Moreover, if the datasets are dynamic (e.g., they are streams of signals from sensors) and change in  quality, then the choice of datasets may have to change dynamically as well. Second, many data discovery tools rely on dataset owners to annotate their datasets for better discovery, but more automatic techniques for understanding and extracting metadata from the data are needed.

While most of the data collection work assumes that the model training comes after the data collection, another important avenue is to augment or improve the data based on how the model performs. While there is a heavy literature on model interpretation~\cite{Ribeiro:2016:WIT:2939672.2939778,anchors:aaai18}, it is not clear how to address feedback on the data level. In the model fairness literature~\cite{Krasanakis:2018:ASR:3178876.3186133}, one approach to reducing unfairness is to fix the data. In data cleaning, ActiveClean and BoostClean are interesting approaches for fixing the data to improve model accuracy. A key challenge is analyzing the model, which becomes harder as models become more complicated.
\customparagraph{Performance Tradeoff}
While traditional labeling techniques focus on accuracy, there is a recent push towards generating large amounts of weak labels. We need to better understand the tradeoffs of accuracy versus scalability to make informed decisions on which approach to use when. For example, simply having more weak labels does not necessarily mean the model's accuracy will eventually reach a perfect accuracy. At some point, it may be worth investing in humans or using transfer learning to make additional improvements. Such decisions can be made through some trial and error, but an interesting question is whether there is a more systematic way to do such evaluations.
\customparagraph{Crowdsourcing }
Despite the many efforts in crowdsourcing, leveraging humans is still a non-trivial task. Dealing with humans involves designing the right tasks and interfaces, ensuring that the worker quality is good enough, and setting the right price for tasks. The recent data programming paradigm introduces a new set of challenges where workers now have to implement labeling functions instead of providing labels themselves. One idea is to improve the quality of such collaborative programming by making the programming of labeling functions drastically easier, say by introducing libraries or templates for programming.
\customparagraph{Empirical comparison of techniques }
Although we showed a flowchart on when to use which techniques, it is far from complete, as many factors are application-specific and can only be determined by looking at the data and application. For example, if the model training can be done with a small number of labels, then we may not have to perform data labeling using crowdsourcing. In addition, the estimated human efforts in labeling and data programming may not follow any theoretical model in practice. For example, humans may find programming for certain applications much more difficult and time-consuming than other applications depending on their expertise. Hence, there needs to be more empirical research on the effectiveness of the techniques.
\customparagraph{Generalizing and integrating techniques }
We observed that many data collection techniques were application or data type specific and were often small parts of a larger research. As machine learning becomes widely used in just about any application, there needs to be more effort in generalizing the techniques to other problems. In data labeling, most of the research effort has been focused on classification tasks and much less on regression tasks. An interesting question is which classification techniques can also be extended to regression. It is also worth exploring if application-specific techniques can be generalized further. For example, the NELL system continuously extracts facts from the Web indefinitely. This idea can possibly be applied to collecting any type of data from any source, although the technical details may differ. Finally, given the variety of techniques for data collection, there needs to be more research on end-to-end solutions that combine techniques for data acquisition, data labeling, and improvements of existing data and models.


\section{Conclusion}

As machine learning becomes more widely used, it becomes more important to acquire large amounts of data and label data, especially for state-of-the-art neural networks. Traditionally, the machine learning, natural language processing, and computer vision communities has contributed to this problem -- primarily on data labeling techniques including semi-supervised learning and active learning. Recently, in the era of Big data, the data management community is also contributing to numerous subproblems in data acquisition, data labeling, and improvement of existing data. In this survey, we have investigated the research landscape of how all these technique complement each other and have provided guidelines on deciding which technique can be used when. Finally, we have discussed  interesting data collection challenges that remain to be addressed. In the future, we expect the integration of Big data and AI to happen not only in data collection, but in all aspects of machine learning.

\section*{Acknowledgments}

This research was supported by the Engineering Research Center Program through the National Research Foundation of Korea (NRF) funded by the Korean Government MSIT (NRF-2018R1A5A1059921), by SK Telecom, and by a Google AI Focused Research Award.

\bibliographystyle{IEEEtran}

\end{document}